\documentclass[10pt,twocolumn,letterpaper]{article}
\usepackage[pagenumbers]{cvpr} 

\newcommand{\mypar}[1]{\vspace{1mm}\noindent\textbf{#1}}
\newcommand{\myparit}[1]{\vspace{1mm}\noindent\emph{#1}}

\def\ourtrain{\textsc{Twin}\xspace}
\def\oureval{\textsc{Fgvqa}\xspace}

\newcommand{\sectionrow}[2]{%
  \rowcolor{gray!15}\multicolumn{#1}{l}{\textit{#2}}\\
}

\newcommand{\gain}[1]{\textsubscript{\textcolor{ForestGreen}{\textbf{+#1}}}}

\newcommand{\loss}[1]{\textsubscript{\textcolor{red}{\textbf{-#1}}}}

\definecolor{incorectcolor}{HTML}{FF0000}
\newcommand{\incorrect}[1]{{\bfseries\color{incorectcolor}#1}} 

\definecolor{correctcolor}{HTML}{008000}
\newcommand{\correct}[1]{{\bfseries\color{correctcolor}#1}} 

\usepackage[most]{tcolorbox}
\newtcolorbox{promptbox}[1][]{
  enhanced,
  colback=gray!5,
  colframe=gray!40,
  boxrule=0.5pt,
  arc=3mm,
  left=5pt, right=5pt, top=5pt, bottom=5pt,
  fonttitle=\bfseries,
  coltitle=black,
  breakable,
  title=#1
}

\usepackage{graphicx}
\usepackage[table]{xcolor}

\def\met{\textsc{Met}\xspace}
\def\ilias{\textsc{Ilias}\xspace}

\def\landmarksshort{\textsc{Landmarks}\xspace}
\def\inquire{\textsc{Inquire}\xspace}
\def\cub{\textsc{Cub}\xspace}

\def\seed{\textsc{Seed}\xspace}
\def\pope{\textsc{Pope}\xspace}
\def\cvbench{\textsc{Cv-Bench}\xspace}
\def\aitwod{\textsc{Ai2d}\xspace}
\def\nlvr{\textsc{Nlvr2}\xspace}
\def\chartqa{\textsc{ChartQA}\xspace}
\def\textvqa{\textsc{TextVQA}\xspace}
\def\realworldqa{\textsc{Realworld}\xspace}
\def\mmmu{\textsc{Mmmu}\xspace}
\def\blink{\textsc{Blink}\xspace}
\def\vlmbias{\textsc{VlmBias}\xspace}

\def\mmlu{\textsc{Mmlu}\xspace}
\def\hellaswag{\textsc{HellaSwag}\xspace}
\def\gsmeightk{\textsc{Gsm8K}\xspace}

\def\cifar{\textsc{Cifar100}\xspace}
\def\pets{\textsc{Pets}\xspace}
\def\sun{\textsc{Sun397}\xspace}

\usepackage{siunitx}
\newcolumntype{N}{S[table-format=2.1, table-space-text-post=\gain{00.0}]}

\makeatletter
\newcommand{\disablemaintoc}{%
  \let\orig@addcontentsline\addcontentsline
  \renewcommand{\addcontentsline}[3]{}%
}

\newcommand{\enablemaintoc}{%
  \let\addcontentsline\orig@addcontentsline
}
\makeatother

\definecolor{cvprblue}{rgb}{0.21,0.49,0.74}
\usepackage[pagebackref,breaklinks,colorlinks,allcolors=cvprblue]{hyperref}
\title{Same or Not? Enhancing Visual Perception in Vision-Language Models}

\author{Damiano Marsili \quad 
Aditya Mehta \quad 
Ryan Y. Lin \quad 
Georgia Gkixoari \smallskip \smallskip\\
California Institute of Technology
}

\begin{document}
\disablemaintoc

\twocolumn[{%
\renewcommand\twocolumn[1][]{#1}%
\maketitle
\vspace{-9mm}
\begin{center}
    \centering
    \includegraphics[width=0.99\linewidth]{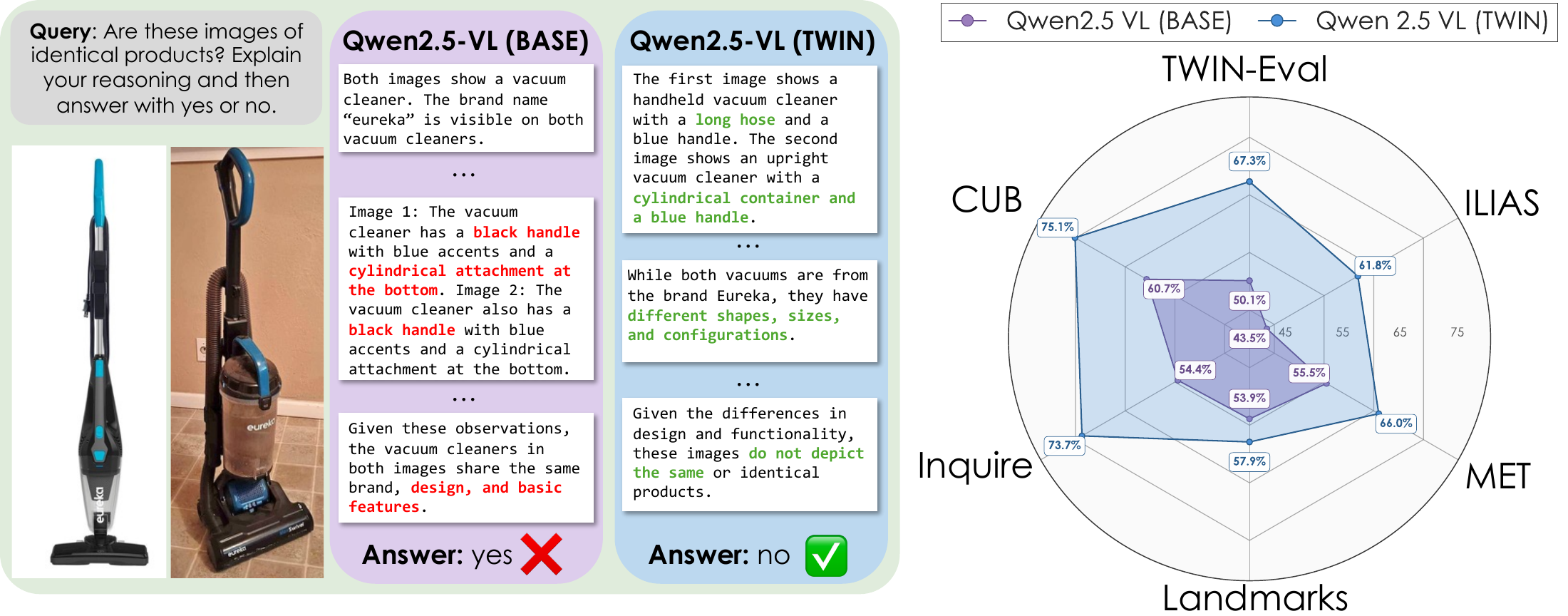}
    \begin{minipage}{0.57\linewidth}
    \centering
    (a) The ~\ourtrain Dataset
    \end{minipage}
    \begin{minipage}{0.42\linewidth}
    \centering
    (b) The ~\oureval Benchmark Suite
    \end{minipage}
\vspace{-2mm}
    \captionof{figure}{\textbf{(a)} Fine-grained visual understanding requires recognizing subtle differences between similar images. We introduce \ourtrain, a large-scale dataset of 561K image-pair queries asking if two images depict the same object, and \oureval, a benchmark suite evaluating fine-grained VQA capabilities. \textbf{(b)} Models trained on \ourtrain~show improved fine-grained understanding on \oureval, even in unseen domains.} 
    \label{fig:teaser}
\end{center}%
}]

\begin{abstract}
Vision–language models (VLMs) excel at broad visual understanding but remain coarse-grained, exhibit visual biases, and miss subtle visual details. Existing training corpora reinforce this limitation by emphasizing general recognition (“Is it a cat or a dog?”) over fine-grained perception. To address this, we introduce a new training corpus and task designed to enhance the perceptual abilities of VLMs. \ourtrain is a large-scale dataset of 561{,}000 image-pair queries that task models to determine whether two visually similar images depict the same object, encouraging attention to nuanced visual cues. The dataset spans a diverse range of everyday objects across contexts, viewpoints, and appearances. Fine-tuning VLMs on \ourtrain yields notable gains in fine-grained recognition, even on unseen domains such as art, animals, plants, and landmarks. To quantify these gains, we introduce \oureval, a benchmark suite of 12{,}000 queries that repurposes fine-grained recognition and retrieval datasets from multiple domains. While existing VLMs struggle on \oureval, when fine-tuned on \ourtrain they improve by up to $19.3\%$, without compromising performance on general VQA benchmarks. Finally, our \ourtrain dataset scales favorably with object annotations, and our analysis shows that scale is key to performance. We envision \ourtrain as a drop-in addition to open-source VLM training corpora, advancing perceptual precision of future models. Project webpage: \href{https://glab-caltech.github.io/twin/}{https://glab-caltech.github.io/twin/}
\end{abstract}
\section{Introduction}
\label{sec:intro}

Fine-grained visual understanding -- recognizing subtle details within images -- is a hallmark of human vision and a desirable goal for machine perception.
Modern vision-language models (VLMs) have shown tremendous progress in broad visual reasoning, however their perception remains coarse-grained: they exhibit systematic biases~\cite{vlind, visioncheckup, vlmbias, bscore} and overlook subtle details~\cite{blink, vlmsblind}. \cref{fig:teaser}(a) showcases these challenges. While the vacuum cleaners share the same color and brand name, they are distinct instances, distinguishable by the geometry of their dustbins, handle, and base. Qwen2.5-VL~\cite{qwen2.5vl}, a strong open-source VLM, incorrectly identifies them as identical and reveals flawed reasoning about their visual attributes.

We attribute the limitations of current VLMs in fine-grained perception partly to their training data. Most large-scale image–text corpora emphasize general visual reasoning -- such as spatial relations~\cite{clevr, marsili2025visual, gqa, cambrian1}, common knowledge~\cite{mmmu, okvqa}, grounding~\cite{visualgenome, flickr30k, dci, refcoco, vcr}, or mathematical reasoning~\cite{mathqa, mathvista, mathverse, math-vision-dataset} -- over detailed visual discrimination. While these datasets enable broad understanding, they provide little incentive to attend to subtle, instance-level differences. This imbalance is stark in leading open-data VLMs~\cite{perceptionlm, smolvlm}, 
where the majority of their training data does not reward visual distinctions beyond category-level understanding.

We ask: what if VLMs were trained on data that rewards fine-grained visual understanding? To explore this question, we introduce \ourtrain, a large-scale VQA dataset of $561{,}000$ instance-centric queries. We name our dataset \ourtrain as it features TWo-image INstance comparisons: each query asks whether two images depict the same physical object instance, requiring attention to subtle visual cues beyond category-level semantics. We source images of household objects across diverse backgrounds, viewpoints, contexts, and lighting conditions, verified by human annotators. We focus on household objects as their images are readily available and they are central to common visual tasks such as robotics and embodied interaction. \cref{fig:teaser}(a) shows a sample from \ourtrain and illustrates the challenge: distinguishing the vacuum cleaners requires noting differences in form factor and shape, rather than shared brand or color.

To evaluate progress in fine-grained visual perception in VLMs, we introduce \oureval, a benchmark suite for fine-grained VQA.
While fine-grained recognition tasks have long existed in vision research, they are typically formulated as 1-of-N classification or retrieval problems, making them ill-suited for VLMs. \oureval adapts and reformulates existing recognition and retrieval datasets for VLM evaluation, comprising $12{,}000$ queries from diverse domains: \met~\cite{met} with artworks such as paintings and sculptures, \inquire~\cite{inquire} featuring animals and plants from iNaturalist~\cite{inaturalist}, \cub~\cite{cub} with bird species, \ilias~\cite{ilias} showcasing retail products, and \landmarksshort~\cite{landmarksv2} containing landmarks such as monuments and architectural structures.

VLMs post-trained on \ourtrain show substantial gains on \oureval, as illustrated in \cref{fig:teaser}(b) which compares the base and post-trained Qwen2.5-VL. Notably, the gains are substantial even on domains that span beyond \ourtrain, \eg in \inquire and \cub. Through qualitative analyses of model explanations and probing experiments on the vision encoder, we find that VLMs trained on \ourtrain attend more effectively to nuanced, distinguishing visual cues. Finally, our scaling analysis shows that scale is crucial for performance, establishing \ourtrain, which scales in size favorably with the number of object annotations, as a promising path for advancing perceptual precision in VLMs.
\section{Related Work}
\label{sec:related}

\mypar{Limitations of VLMs.} VLMs trained on large text-image corpora have shown remarkable capabilities in image understanding. Here, proprietary systems~\cite{gpt-5, gpt4, gemini2.5, claude} lead in performance, while open-source alternatives~\cite{vila, deepseekvl2, llama3, molmo, moondream2, smolvlm, gemma} are narrowing the gap. Despite these advances, VLMs exhibit systematic biases~\cite{vlmbias, vlind, bscore, visioncheckup}, struggle with fine-grained visual perception~\cite{blink, vlmsblind}, and often hallucinate details~\cite{pope, eyeswideshut, autohallusion, beaf}. To address these challenges, we introduce a large-scale VQA dataset to enhance fine-grained visual understanding in VLMs along with a benchmark suite to evaluate this capability.

\mypar{VLM Training Data.} While most proprietary VLMs rely on closed-source training data~\cite{gpt-5, gpt4, claude, gemini2.5, gemini}, recent efforts have prioritized open-data alternatives. Molmo~\cite{molmo} introduced an open-data model competitive with proprietary systems, while SmolVLM~\cite{smolvlm} and PerceptionLM~\cite{perceptionlm} scale this paradigm using diverse datasets spanning captioning~\cite{coco, nocaps, young2014image, docci, altogether}, grounding~\cite{visualgenome, flickr30k, dci, refcoco, vcr}, spatial reasoning~\cite{blink, clevr, vsr}, math~\cite{intergps, gsm8k, mathqa}, and chart/diagram understanding~\cite{chartqa, charttotext, ai2d, vistext, charxiv}. Most relevant to us are SpotTheDiff~\cite{spotthediff} and Birds-to-Words~\cite{birdstowords}, which pair images with text describing differences. In contrast, \ourtrain is more comprehensive and larger in size: \ourtrain features both matching and non-matching pairs, enabling reasoning about sameness and difference, distinctions in \ourtrain are finer and pertain to subtle cues, \ourtrain is far larger ($561$K pairs vs.\ $13$K of SpotTheDiff and $3.3$K of Birds-to-Words), and it spans diverse object types beyond a single domain such as birds. A detailed comparison is in the Appendix.

\mypar{Fine-grained Recognition} categorizes images into specific subgroups~\cite{finegrainedsurvey}, with applications in classifying plants~\cite{inaturalist,oxfordflowers, inquire}, birds~\cite{cub2011, nabirds}, vehicles~\cite{fgvcaircrafts, stanfordcars} and recently, retail products~\cite{imaterialist, products10k, aliproducts, ilias}. These tasks are typically framed as 1-of-N classification or retrieval, making them ill-suited for evaluating VLMs. We repurpose popular benchmarks~\cite{ilias, met, inquire,cub, landmarksv2} and design a task to evaluate the fine-grained VQA abilities of VLMs.

\mypar{VLM Post-training.} Post-training refines VLMs by instilling specialized capabilities and aligning them with human intent. Supervised fine-tuning (SFT)~\cite{star, s1, vocot} and distillation~\cite{llava-cot, visprogdistill, ical} transfer task-specific skills from annotated data or teacher models, but often cause forgetting and reduced generalization~\cite{rlsrazor, sftmemorizesrlgeneralizes}. Reinforcement learning mitigates these issues, advancing instruction-following~\cite{rlhf, dpo, qwen3}, reasoning~\cite{deepseek-r1, deepseekvl2, drgrpo, qwen3, gemini2.5}, and spatial understanding~\cite{vigorl, grit, marsili2025no}. Beyond post-training methodology, the choice of data plays a critical role in driving post-training success.
We demonstrate that reinforcement learning on our fine-grained \ourtrain dataset increases the perceptual precision of VLMs while preserving their performance on core image–language tasks.
\begin{figure*}[t]
    \centering
    \begin{subfigure}[b]{0.60\textwidth}
        \centering
        \includegraphics[width=\linewidth]{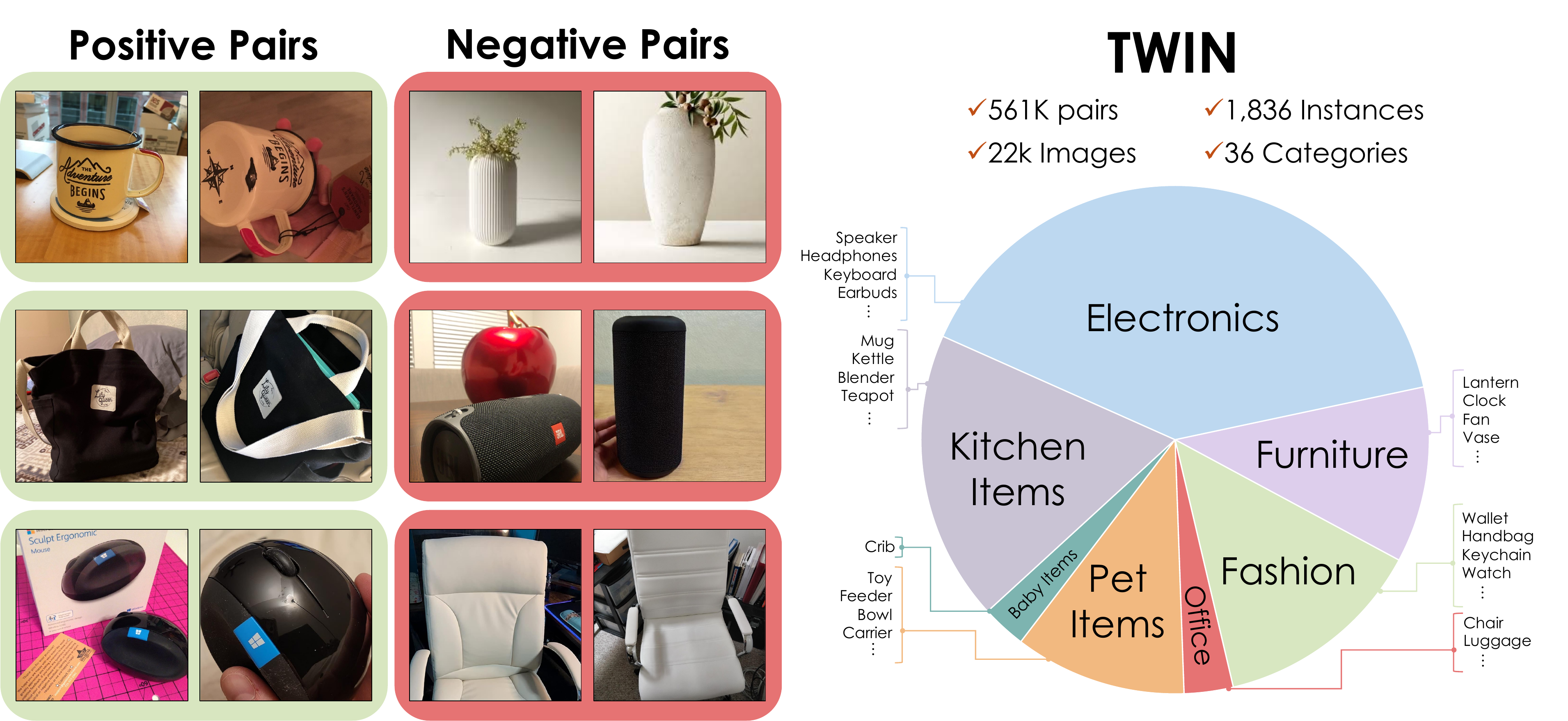}
        \caption{}
        \label{fig:ourdata}
    \end{subfigure}
    \hfill
    \begin{subfigure}[b]{0.38\textwidth}
        \centering
        \includegraphics[width=\linewidth]{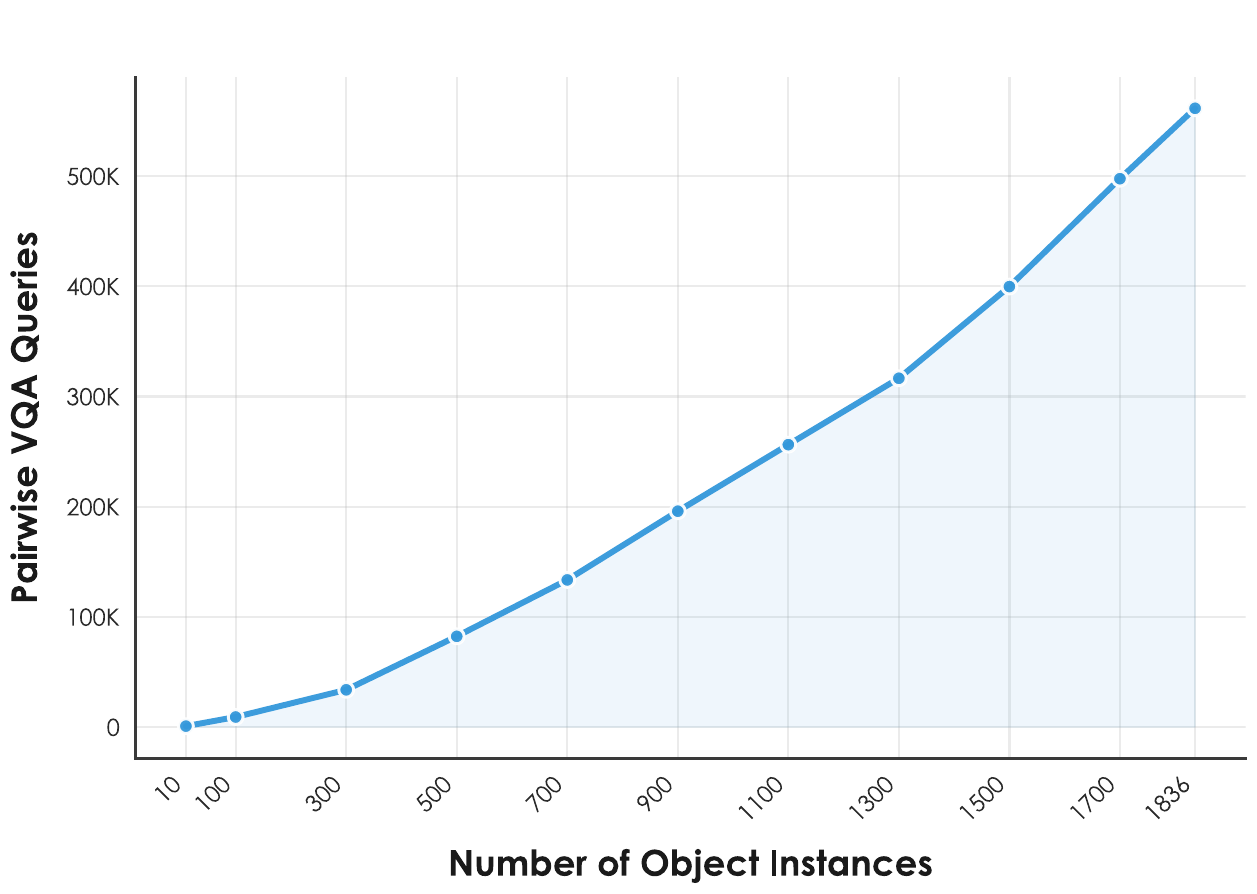}
        \caption{}
        \label{fig:pair_scaling}
    \end{subfigure}
    \vspace{-3mm}
    \caption{(a) \textbf{\ourtrain} is a large-scale VQA dataset for fine-grained visual understanding, where VLMs determine whether two images depict the same instance. \ourtrain contains $561$K pairwise VQA queries across $1,836$ object instances, spanning $36$ categories of common objects and over $22$K images. (b) Scalability of the pairwise construction: from $1,836$ instances, \ourtrain yields $561$K pairwise queries.}
    \label{fig:overall_data}
\end{figure*}
 
\section{The \ourtrain Dataset}
\label{sec:dataset}

We present \ourtrain, a large-scale VQA dataset for advancing fine-grained visual understanding in VLMs. Existing training corpora emphasize broad visual understanding -- spatial relations, common knowledge, and category-level recognition -- offering little supervision for fine perception. As shown in~\cref{fig:teaser}(a), current VLMs often miss subtle cues essential for reliable perception. To address this, \ourtrain introduces $561{,}000$ instance-centric queries where models are tasked to judge whether two similar looking images depict the same object instance. This design rewards attention to nuanced, instance-level details such as shape, texture, and part geometry, going beyond category-level understanding. ~\cref{fig:overall_data} shows samples from \ourtrain and its statistics.

\subsection{Dataset Construction}

\mypar{Task Definition.} In our task, fine-grained visual understanding requires distinguishing object instances -- recognizing when two images show the same object rather than the same category. We define an instance as a set of images of the same physical object under varied viewpoints, lighting, and backgrounds, consistent with prior work~\cite{ilias, met, scalablerecognitionvocabtree, landmarksv2, image-retrieval-deep-local}. From this definition, we design a VQA task where a VLM receives two images and determines if they depict the same instance. Unlike category-level recognition (\eg, cat vs dog), our task demands sensitivity to fine, instance-specific cues -- shape, texture, and part geometry -- that distinguish similar objects. \cref{fig:teaser}(a) shows the setup; VQA prompts are in the Appendix.

\begin{figure*}[t]
    \centering
    \includegraphics[width=0.95\linewidth]{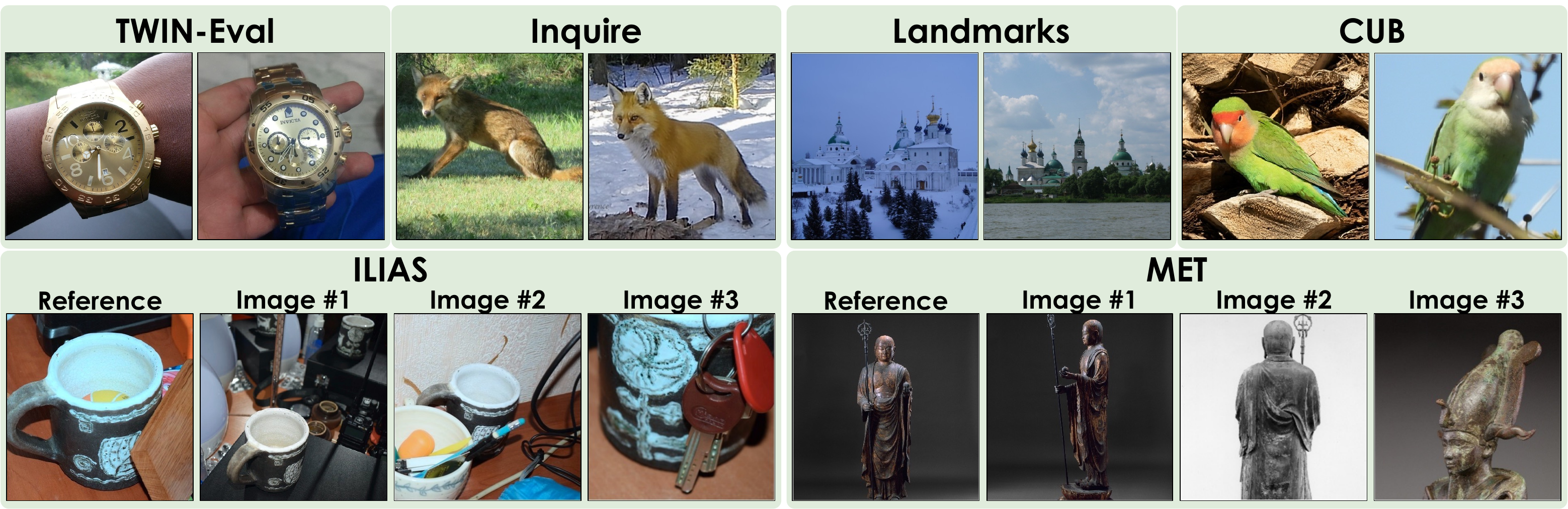}
    \vspace{-3mm}
    \caption{\textbf{\oureval} is a suite of fine-grained VQA benchmarks spanning retail products (\ourtrain, \ilias~\cite{ilias}); animals and plants (\inquire~\cite{inquire}); \landmarksshort~\cite{landmarksv2}; birds (\cub~\cite{cub}); and art (\met~\cite{met}). We include two query types: \emph{pair} (top row), where a VLM judges if two images depict the same instance, species, or landmark, and \emph{multi} where it counts how many images match a reference.}
    \vspace{-4mm}
    \label{fig:fgvqa}
\end{figure*}

\mypar{Sourcing Instances.} We source object instances across diverse categories from Amazon Reviews~\cite{amazon-reviews}. For each object instance, we collect images with varied contexts, viewpoints, and lighting to ensure visual diversity. Human annotators verify that each image clearly shows and identifies the object. We generate positive pairs by sampling from images of the same instance.~\cref{fig:ourdata} shows examples of positive pairs along with the dataset's distribution.

\mypar{Hard Negative Pairs.} A balanced dataset for fine-grained understanding requires both positive and negative pairs. If all examples were positive, the task would be trivial. Likewise, random negatives are often too easy (\eg, a mug paired with a fan). We therefore focus on hard negatives -- distinct objects that appear similar. We collect these hard negatives with the help of human annotators. First, using CLIP~\cite{clip}, we compute pairwise cosine similarities between image embeddings to shortlist visually similar candidates, from which human annotators select the final pairs. Examples of such hard negative pairs are shown in \cref{fig:ourdata}, including two white vases with slightly different shapes and two white chairs differing in texture and armrest design.

\mypar{Generated Negatives.} Collecting hard negatives is costly and difficult to scale, so we augment \ourtrain with synthetic negatives using personalized image generation, DreamBooth~\cite{dreambooth}. Given images of an instance, DreamBooth produces similar variants that retain overall appearance but alter fine details, simulating hard negatives without extra data collection. The rightmost speaker in~\cref{fig:ourdata} demonstrates this: its color and geometry match the left one, but it omits the red logo and lighter contours. Each synthetic image is verified by human annotators. Details on image generation and additional examples are in the Appendix. We validate the impact of these generated negatives in our experiments.

\mypar{Dataset Statistics.} \ourtrain features $561$K pairwise VQA queries, with $123$K positive and $438$K negative samples, spanning $22{,}157$ unique images of $1{,}836$ object instances, with $5{,}288$ generated images. We plot the number of pairwise queries vs the number of object instances in~\cref{fig:pair_scaling}, highlighting the scalability of our pairwise formulation: \ourtrain's size scales favorably with the number of object instances. Our scaling analysis in~\cref{subsec:ablations} demonstrates that collecting \ourtrain at scale is crucial for performance.

\subsection{Training VLMs on \textbf{\ourtrain}}
\begin{figure}[t!]
    \centering
    \includegraphics[width=0.99\columnwidth]{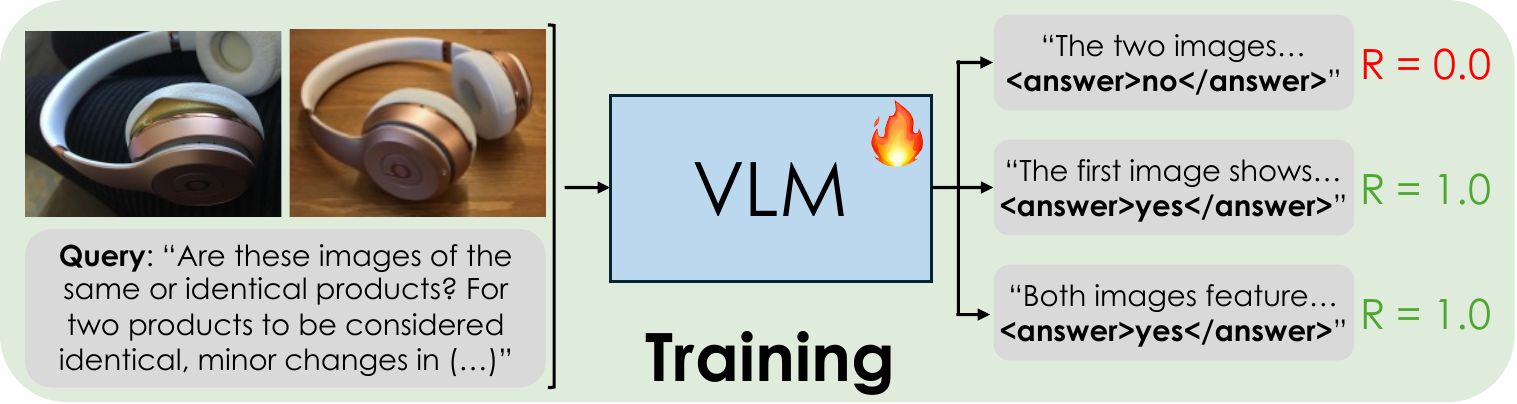}
    \vspace{-2mm}
    \caption{\textbf{Training VLMs on \ourtrain.} We train VLMs using RL on \ourtrain. Reward is computed by comparing the predicted answer with the ground truth pair assignment.}
    \label{fig:model-training}
    \vspace{-6mm}
\end{figure}
To improve fine-grained understanding in VLMs and evaluate the impact of our new dataset,
we post-train existing models on \ourtrain. Our task requires recognizing subtle attributes to distinguish similar instances, and we hypothesize that optimizing for this task enhances broader fine-grained abilities. We post-train with RL as it is shown to improve model capabilities while preserving prior skills~\cite{sftmemorizesrlgeneralizes, rlsrazor, huan2025does, unveilingcompositiongap}. \cref{fig:model-training} outlines our model training approach.

\mypar{Setup.} Given image pairs $(I_1, I_2)$ with ground truth label $y \in \{\text{yes, no}\}$, a VLM $\pi_\theta$, parametrized by $\theta$, is prompted to produce a textual explanation and a final answer $\hat{y}$ whether both images depict the same instance. We use a binary outcome reward comparing prediction and ground truth: $R(y, \hat{y}) =\mathbf{1}_{\{y=\hat{y}\}}$. Importantly, supervision relies \emph{only} on pairwise assignments, without any descriptive textual annotations. 

\mypar{Optimization.} We tune our VLM $\pi_\theta$ using GRPO~\cite{deepseekmath} as it balances maximizing expected advantages while preventing drift from the pre-trained VLM. Details are in the Appendix.

\mypar{Implementation.} We train Qwen2.5-VL-3B-Instruct~\cite{qwen2.5vl} and InternVL3.5-1B-Instruct~\cite{internvl3} on \ourtrain, chosen as leading open-source VLMs at the 3B and 1B scales. Training is end-to-end on $4$ A100 GPUs for $1$ epoch with a batch size of $480$, group size $5$, and learning rate $10^{-6}$. We use verl~\cite{verl} for optimization and balance positive and negative queries in each batch. Hyperparameters are in the Appendix.
\section{The \oureval Benchmark Suite}
\label{sec:fgvqa}

Despite their strong general visual abilities, current VLMs struggle with fine-grained understanding~\cite{vlmbias, vlmsblind}. Existing VQA benchmarks emphasize broad image understanding, leaving this capability largely unevaluated. To this end, we introduce \oureval, a suite of fine-grained VQA benchmarks that measure how well VLMs identify subtle visual cues. We show examples in~\cref{fig:fgvqa} and provide more details below.

\mypar{Benchmarks.} Fine-grained understanding is a general skill -- models attuned to subtle differences should generalize across domains. To evaluate this, we repurpose recognition and retrieval datasets, totaling $12{,}000$ queries spanning diverse domains: \met~\cite{met} (artworks such as paintings and sculptures), \ilias~\cite{ilias} (retail products), \landmarksshort~\cite{landmarksv2} (monuments), \cub~\cite{cub} (bird species), \inquire~\cite{inquire} (animals and plants from iNaturalist~\cite{inaturalist}), and \ourtrain-Eval, collected like \ourtrain but with unseen instances.
The breadth of \oureval enables assessment of cross-domain generalization.

\begin{table*}[]
\resizebox{0.99\linewidth}{!}{
\begin{tabular}{l|N|NNN|NNN|NNN|NNN|NNN|NNN}
\toprule
 & & \multicolumn{3}{c|}{\ourtrain-Eval} & \multicolumn{3}{c|}{\ilias~\cite{ilias}} & \multicolumn{3}{c|}{\landmarksshort~\cite{landmarksv2}} & \multicolumn{3}{c|}{\met~\cite{met}} & \multicolumn{3}{c|}{\cub~\cite{cub}} & \multicolumn{3}{c}{\inquire~\cite{inquire}} \\
 & \multicolumn{1}{l|}{\textsc{Mean}} & \multicolumn{1}{l}{Pair} & \multicolumn{1}{l}{Multi} & \multicolumn{1}{l|}{Total}
 & \multicolumn{1}{l}{Pair} & \multicolumn{1}{l}{Multi} & \multicolumn{1}{l|}{Total}
 & \multicolumn{1}{l}{Pair} & \multicolumn{1}{l}{Multi} & \multicolumn{1}{l|}{Total}
 & \multicolumn{1}{l}{Pair} & \multicolumn{1}{l}{Multi} & \multicolumn{1}{l|}{Total}
 & \multicolumn{1}{l}{Pair} & \multicolumn{1}{l}{Multi} & \multicolumn{1}{l|}{Total}
 & \multicolumn{1}{l}{Pair} & \multicolumn{1}{l}{Multi} & \multicolumn{1}{l}{Total} \\
\midrule

\sectionrow{20}{\textcolor{gray!70}{Proprietary VLMs}}
{\textcolor{gray!70}{GPT4o}~\cite{gpt4}} &
\textcolor{gray!70}{\tablenum{86.1}} &
\textcolor{gray!70}{\tablenum{92.2}} & \textcolor{gray!70}{\tablenum{71.2}} & \textcolor{gray!70}{\tablenum{81.7}} &
\textcolor{gray!70}{\tablenum{95.0}} & \textcolor{gray!70}{\tablenum{95.5}} & \textcolor{gray!70}{\tablenum{95.3}} &
\textcolor{gray!70}{\tablenum{87.3}} & \textcolor{gray!70}{\tablenum{74.7}} & \textcolor{gray!70}{\tablenum{81.0}} &
\textcolor{gray!70}{\tablenum{86.3}} & \textcolor{gray!70}{\tablenum{72.9}} & \textcolor{gray!70}{\tablenum{79.6}} &
\textcolor{gray!70}{\tablenum{91.7}} & \textcolor{gray!70}{\tablenum{86.0}} & \textcolor{gray!70}{\tablenum{88.9}} &
\textcolor{gray!70}{\tablenum{94.3}} & \textcolor{gray!70}{\tablenum{86.3}} & \textcolor{gray!70}{\tablenum{90.3}} \\

{\textcolor{gray!70}{Gemini 2.5 Flash}~\cite{gemini2.5}} &
\textcolor{gray!70}{\tablenum{82.2}}
&
\textcolor{gray!70}{\tablenum{89.1}} & \textcolor{gray!70}{\tablenum{68.1}} & \textcolor{gray!70}{\tablenum{78.6}} &
\textcolor{gray!70}{\tablenum{92.2}} & \textcolor{gray!70}{\tablenum{90.5}} & \textcolor{gray!70}{\tablenum{91.4}} &
\textcolor{gray!70}{\tablenum{85.4}} & \textcolor{gray!70}{\tablenum{77.0}} & \textcolor{gray!70}{\tablenum{81.2}} &
\textcolor{gray!70}{\tablenum{82.6}} & \textcolor{gray!70}{\tablenum{71.9}} & \textcolor{gray!70}{\tablenum{77.3}} &
\textcolor{gray!70}{\tablenum{84.1}} & \textcolor{gray!70}{\tablenum{77.9}} & \textcolor{gray!70}{\tablenum{81.0}} &
\textcolor{gray!70}{\tablenum{88.3}} & \textcolor{gray!70}{\tablenum{79.4}} & \textcolor{gray!70}{\tablenum{83.9}} \\

\midrule
\sectionrow{20}{Open-Source VLMs}

PerceptionLM 3B~\cite{perceptionlm} & 37.8
& 46.9 & 25.2 & 36.1 & 49.6 & 25.0 & 37.3 & 51.5 & 25.7 & 38.6 &
50.4 & 26.1 & 38.3 & 49.6 & 24.9 & 37.3 & 52.1 & 26.1 & 39.1 \\

Moondream 2 2B~\cite{moondream2} & 45.9 & 
64.5 & 30.0 & 47.3 & 58.7 & 30.3 & 44.5 & 59.8 & 25.2 & 42.5 &
84.3 & 27.0 & 55.7 & 50.6 & 43.3 & 47.0 & 51.9 & 25.1 & 38.5 \\

Gemma3 4B~\cite{gemma3} & 68.7 & 
68.1 & 39.5 & 53.8 & 88.4 & 46.5 & 67.5 & 91.2 & 55.2 & 73.2 &
87.9 & 57.5 & 72.7 & 91.2 & 55.2 & 73.2 & 90.0 & 48.7 & 69.4 \\

SmolVLM2 2.2B~\cite{smolvlm} & 46.9 &
54.7 & 25.1 & 39.9 & 52.2 & 29.7 & 41.0 & 54.1 & 29.6 & 41.9 &
73.9 & 27.9 & 50.9 & 81.5 & 35.0 & 58.3 & 73.0 & 25.7 & 49.4 \\

\midrule
\sectionrow{20}{Direct Comparisons}

Qwen2.5-VL 3B~\cite{qwen2.5vl} & 53.0 & 
70.4 & 29.8 & 50.1 & 54.5 & 32.5 & 43.5 & 79.5 & 28.2 & 53.9 &
81.6 & 29.4 & 55.5 & 84.0 & 37.4 & 60.7 & 74.4 & 34.3 & 54.4 \\

+ \ourtrain & {\bfseries 67.0}\gain{14.0} & 
{\bfseries 88.9}\gain{18.5} & {\bfseries 45.7}\gain{15.9} & {\bfseries 67.3}\gain{17.2} &
{\bfseries 79.9}\gain{25.4} & {\bfseries 43.7}\gain{11.2} & {\bfseries 61.8}\gain{18.3} &
{\bfseries 80.8}\gain{1.3}  & {\bfseries 34.9}\gain{6.7}  & {\bfseries 57.9}\gain{4.0} &
{\bfseries 85.4}\gain{3.8}  & {\bfseries 46.5}\gain{17.1} & {\bfseries 66.0}\gain{10.5} &
{\bfseries 92.5}\gain{8.5}  & {\bfseries 57.7}\gain{20.3} & {\bfseries 75.1}\gain{14.4} &
{\bfseries 92.7}\gain{18.3} & {\bfseries 54.6}\gain{20.3} & {\bfseries 73.7}\gain{19.3} \\

\midrule

InternVL3.5 1B~\cite{internvl3} & 51.3 &
67.2 & 25.9 & 46.6 & 60.0 & {\bfseries \tablenum{39.1}} & 49.6 &
54.1 & 33.5 & 43.8 & 73.9 & 47.1 & 60.5 & 56.1 & 40.5 & 48.3 &
{\bfseries \tablenum{80.4}} & 37.9 & {\bfseries \tablenum{59.2}} \\
+ \ourtrain & {\bfseries 54.2}\gain{2.9} & 
{\bfseries 73.0}\gain{5.8} & {\bfseries 26.5}\gain{0.6} & {\bfseries 49.8}\gain{3.2} &
{\bfseries 73.2}\gain{13.2} & \tablenum{38.0\loss{1.1}} & {\bfseries 55.6}\gain{6.0} &
{\bfseries 57.2}\gain{3.1} & {\bfseries 36.3}\gain{2.8} & {\bfseries 46.8}\gain{3.0} &
{\bfseries 77.6}\gain{3.7} & {\bfseries 47.5}\gain{0.4} & {\bfseries 62.6}\gain{2.1} &
{\bfseries 62.6}\gain{6.5} & {\bfseries 41.5}\gain{1.0} & {\bfseries 52.1}\gain{3.8} &
\tablenum{76.0\loss{-4.4}} & {\bfseries 40.0}\gain{2.1} & \tablenum{58.0\loss{1.2}} \\

\bottomrule
\end{tabular}
}
\vspace{-2mm}
\caption{
\textbf{Accuracy (\%) on~\oureval.} For \emph{Pair} queries, VLMs are shown two images and asked if they show the same instance. On \emph{multi} queries, VLMs are shown a reference image and three candidates and asked how many show the same instance as the reference. We report accuracy on each constituent dataset separately and the mean. Training on~\ourtrain~improves performance, even on domains not in~\ourtrain.}
\label{table:main_results}
\vspace{-4mm}
\end{table*}

\begin{figure*}[t!!]
    \centering
    \includegraphics[width=0.49\linewidth]{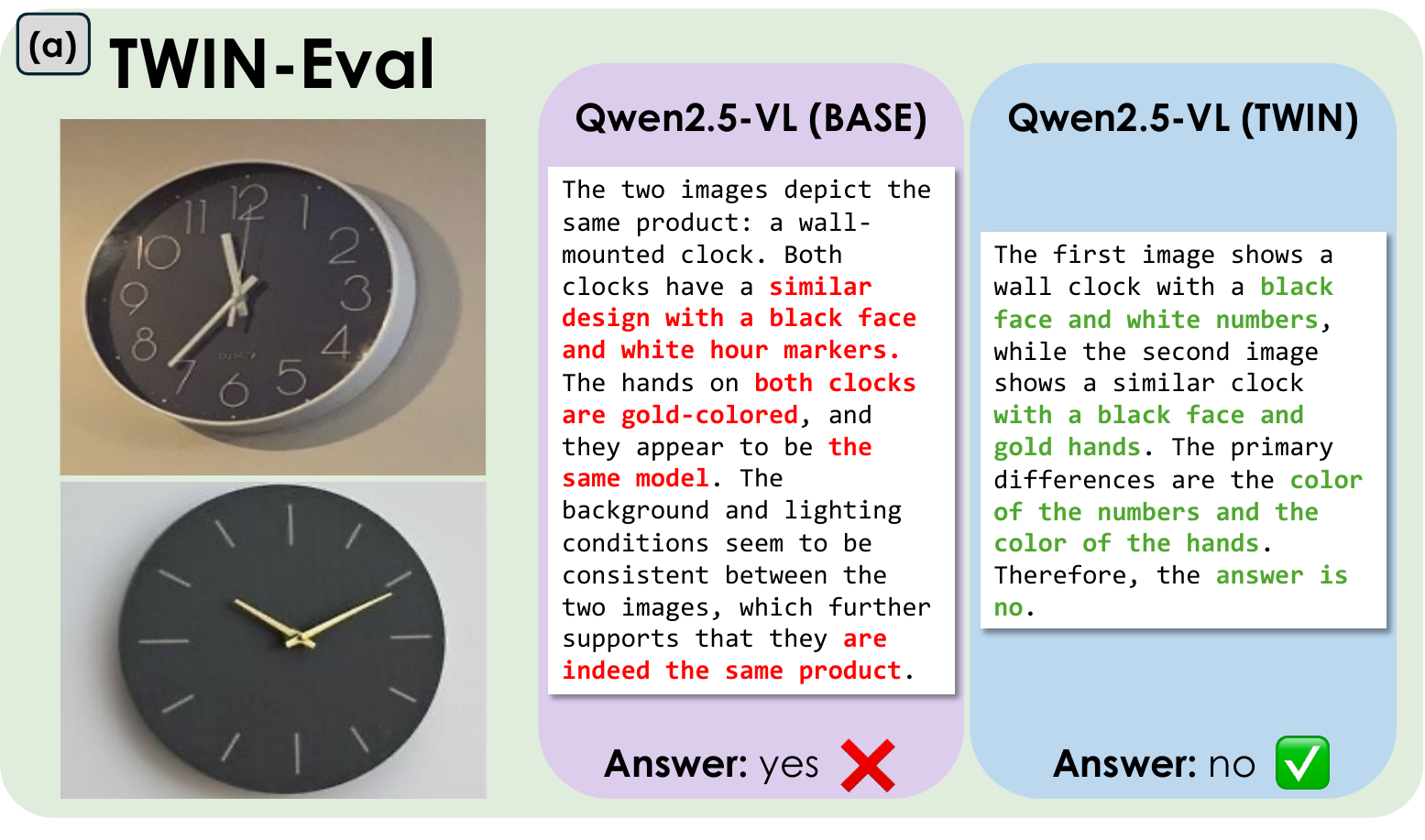}
    \vspace{0.5mm}
    \includegraphics[width=0.49\linewidth]{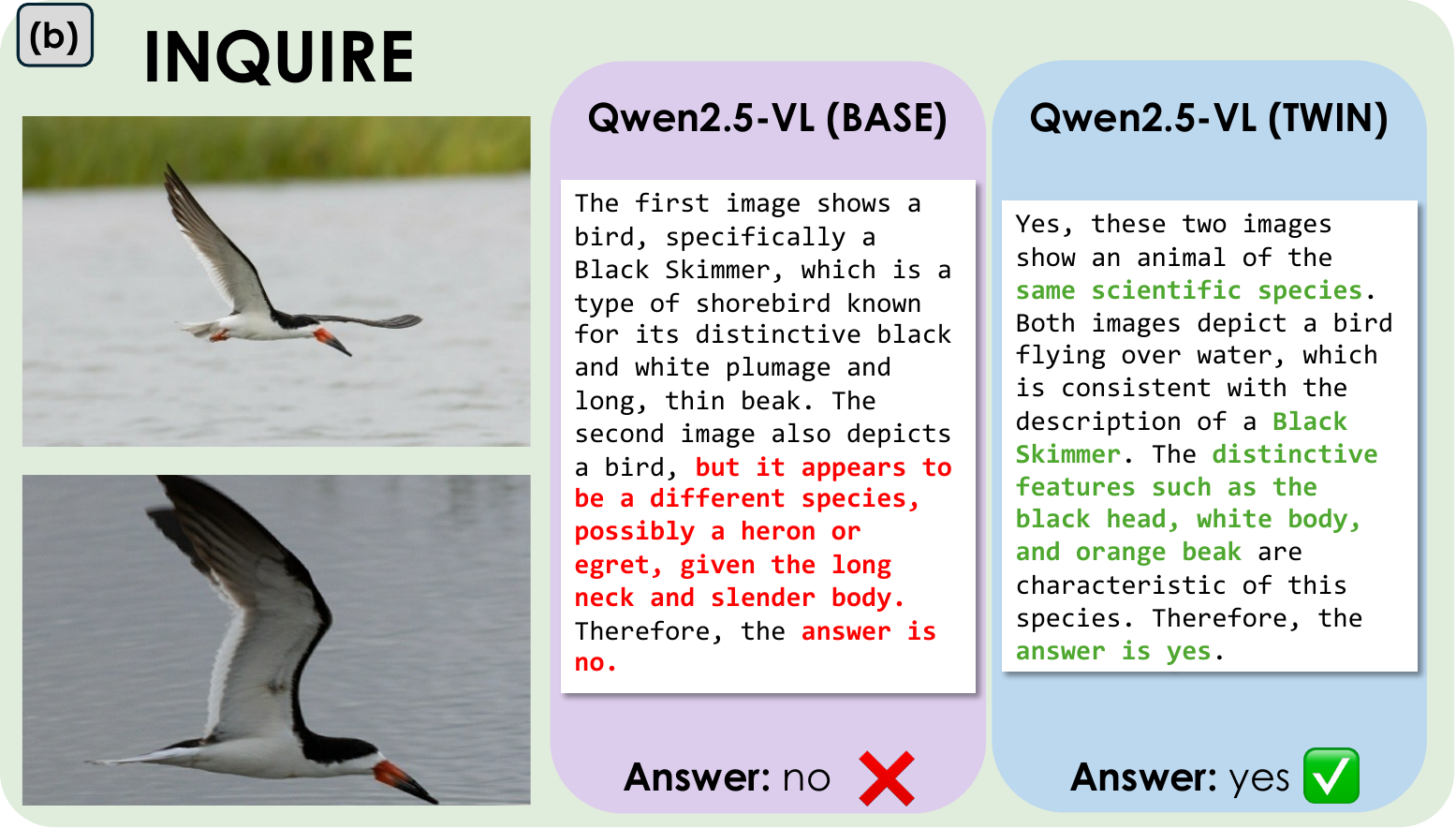}
    \vspace{0.5mm}
    \includegraphics[width=0.49\linewidth]{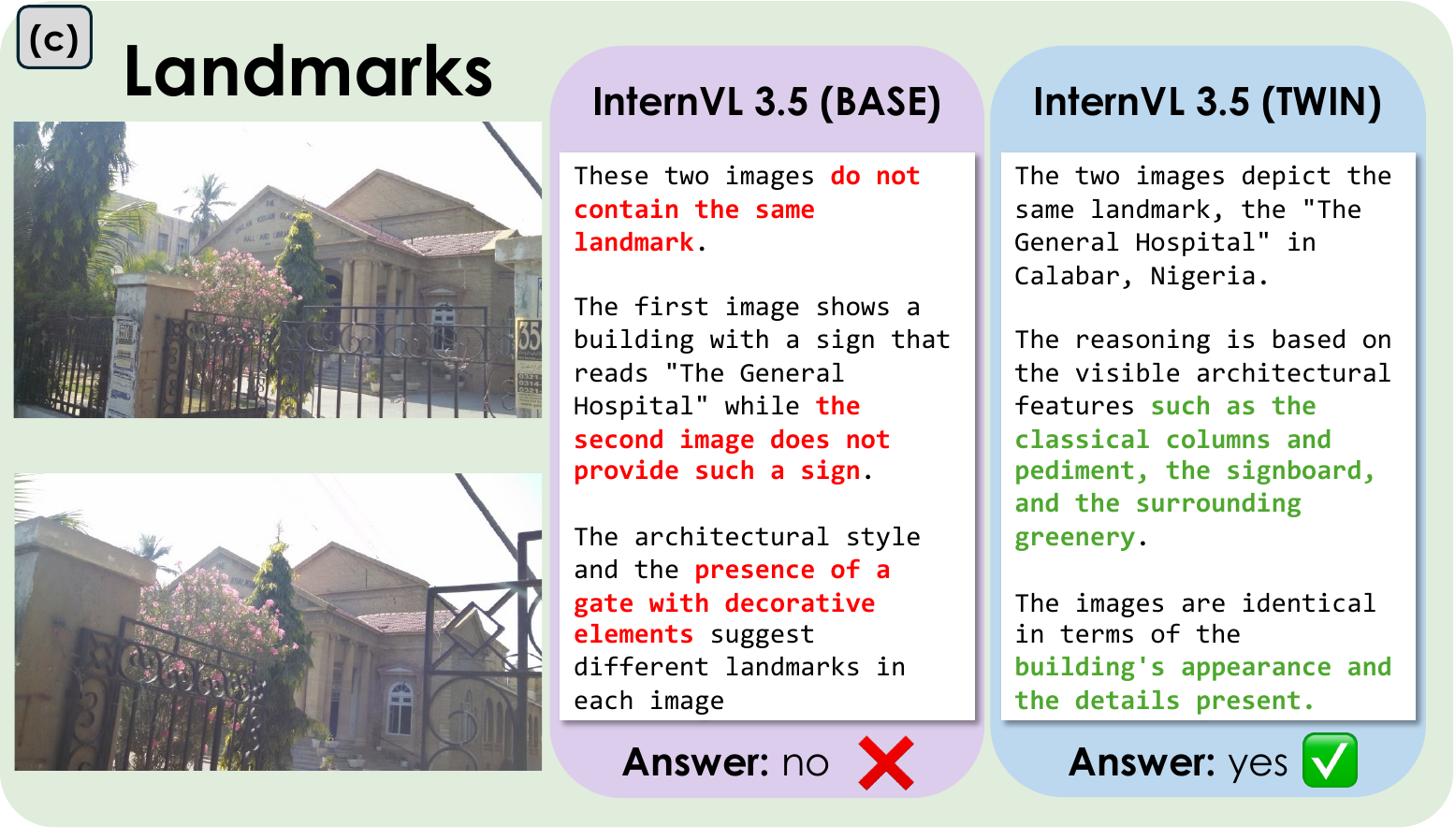}
    \includegraphics[width=0.49\linewidth]{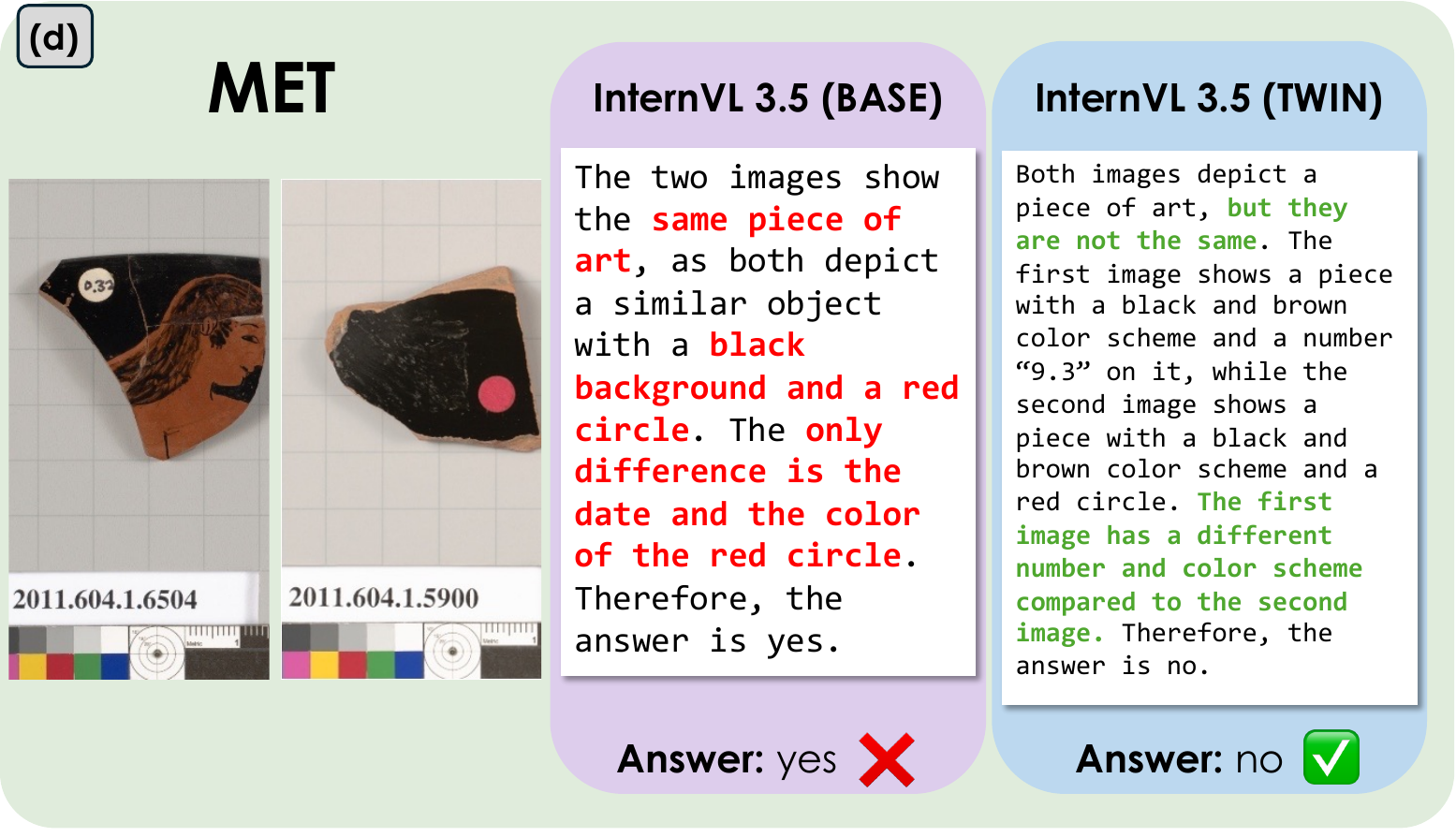}
    \vspace{0.5mm}
    \includegraphics[width=0.49\linewidth]{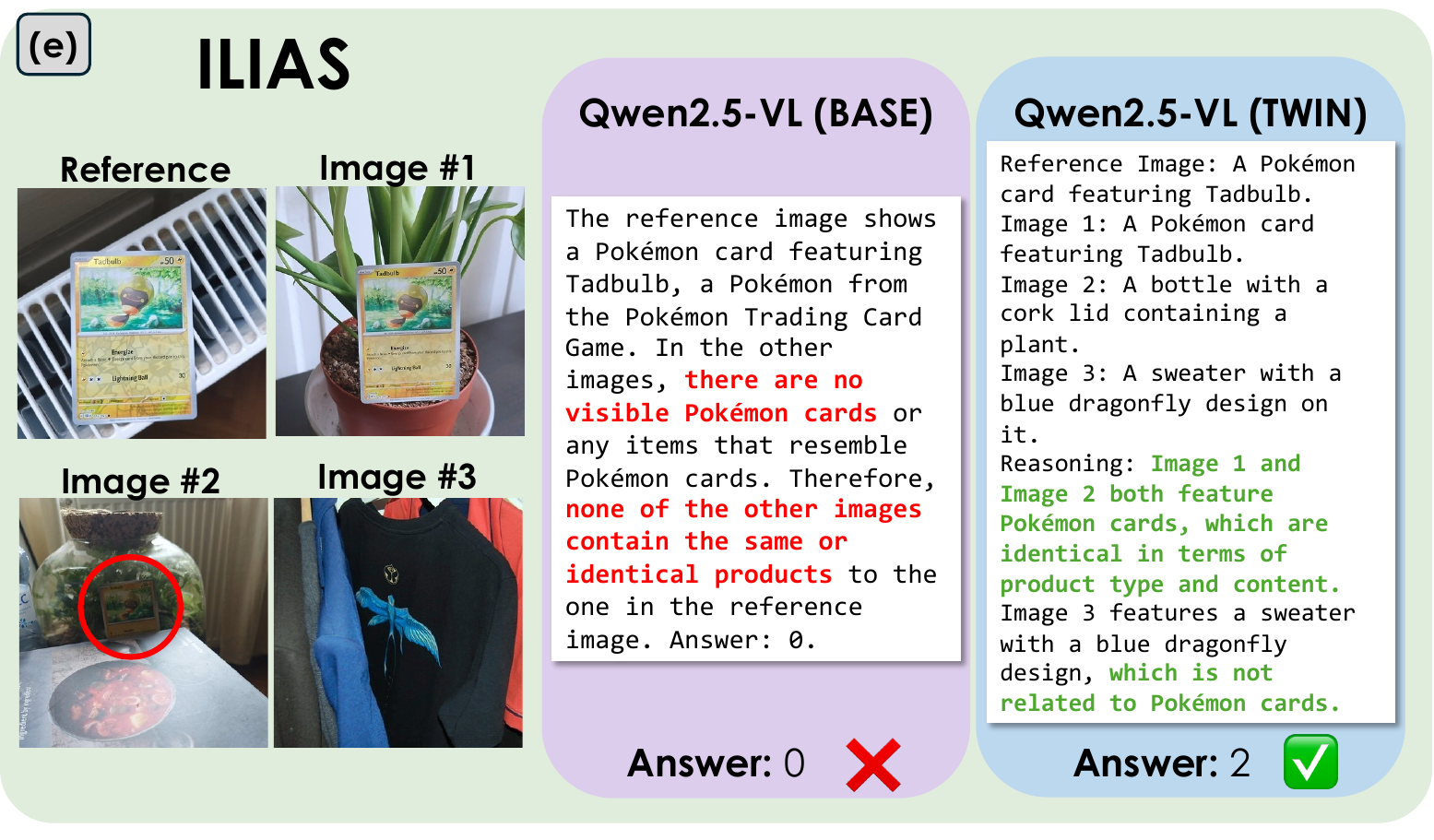}
    \includegraphics[width=0.49\linewidth]{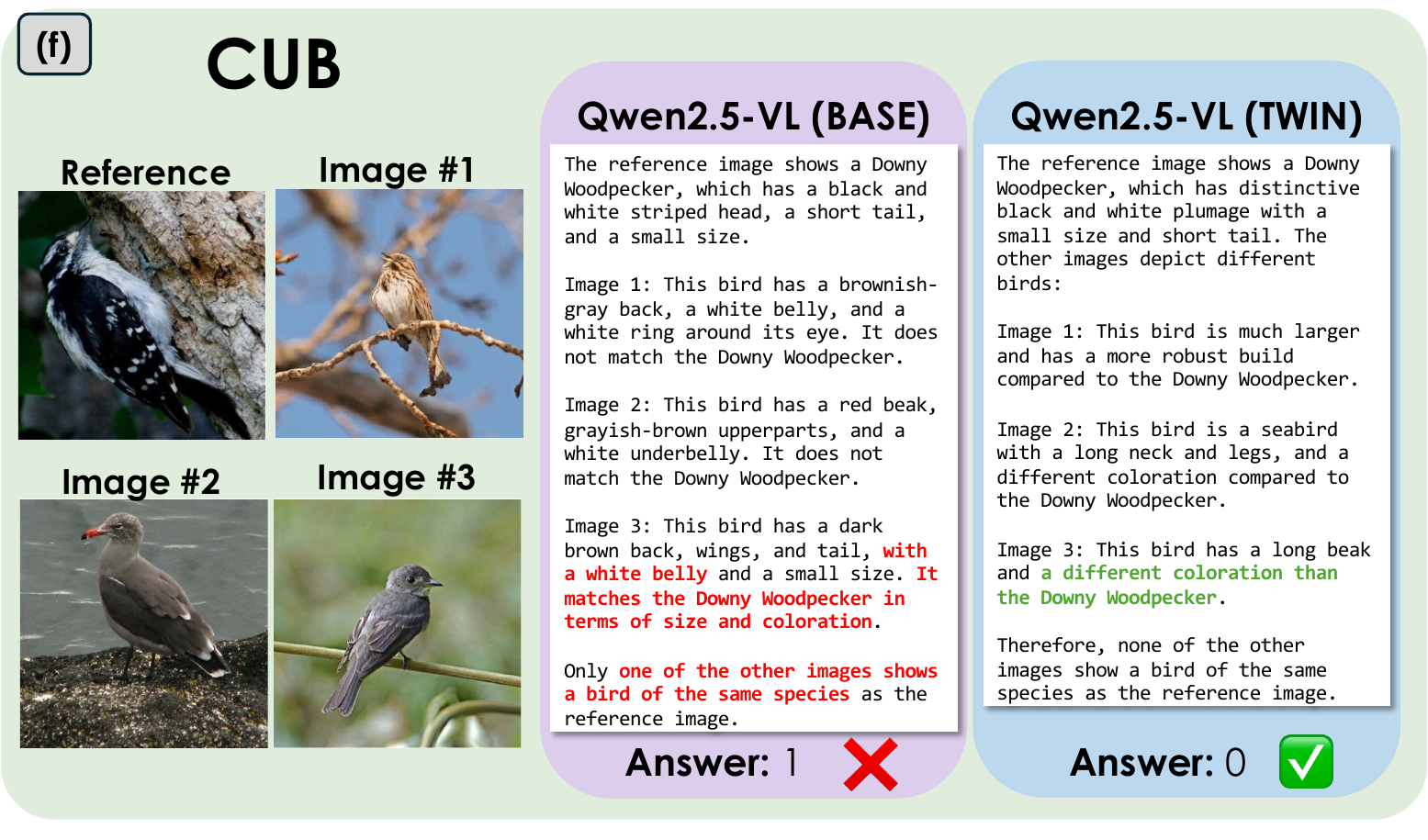}
    \vspace{0.5mm}
    \includegraphics[width=0.49\linewidth]{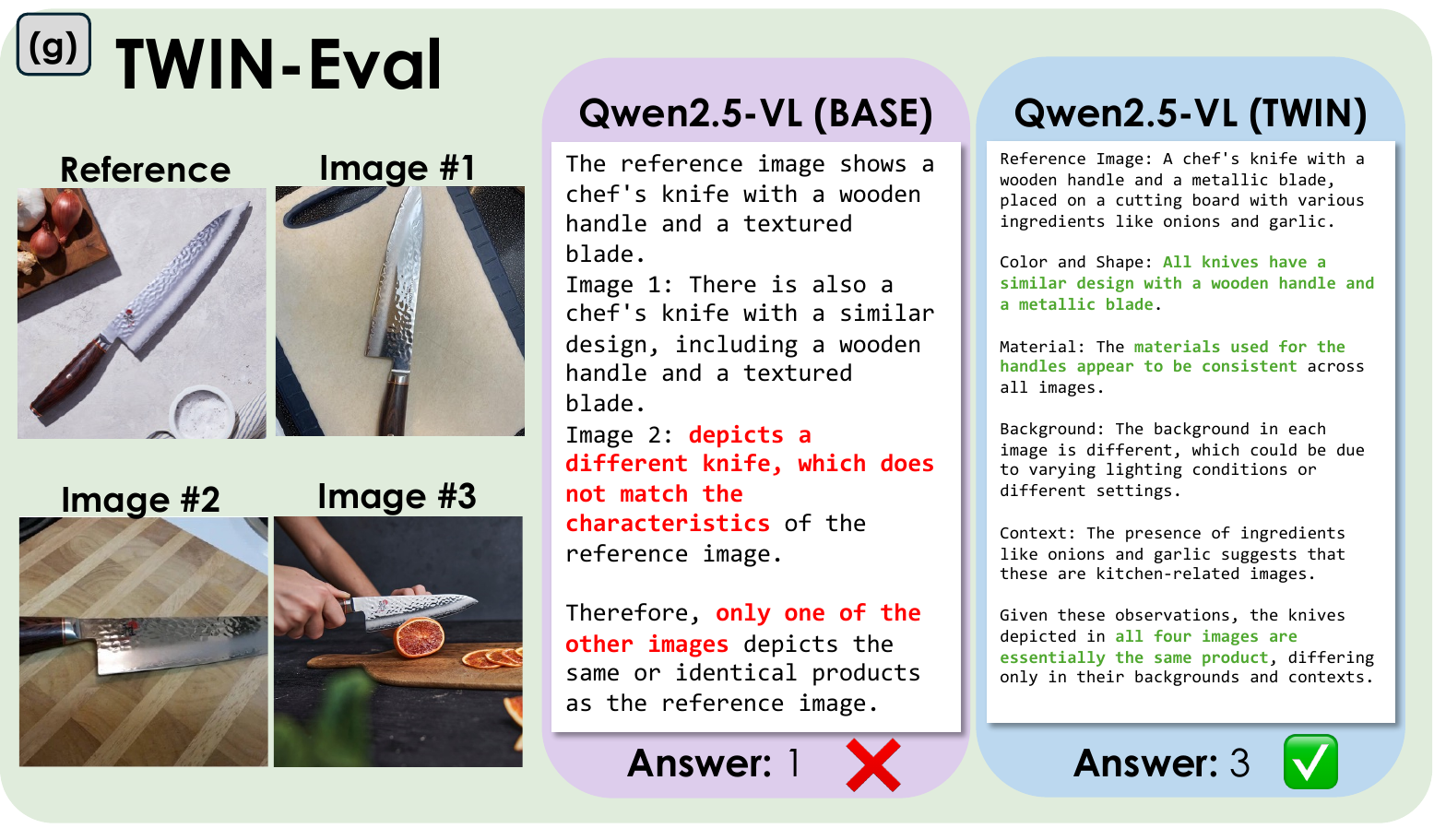}
    \includegraphics[width=0.49\linewidth]{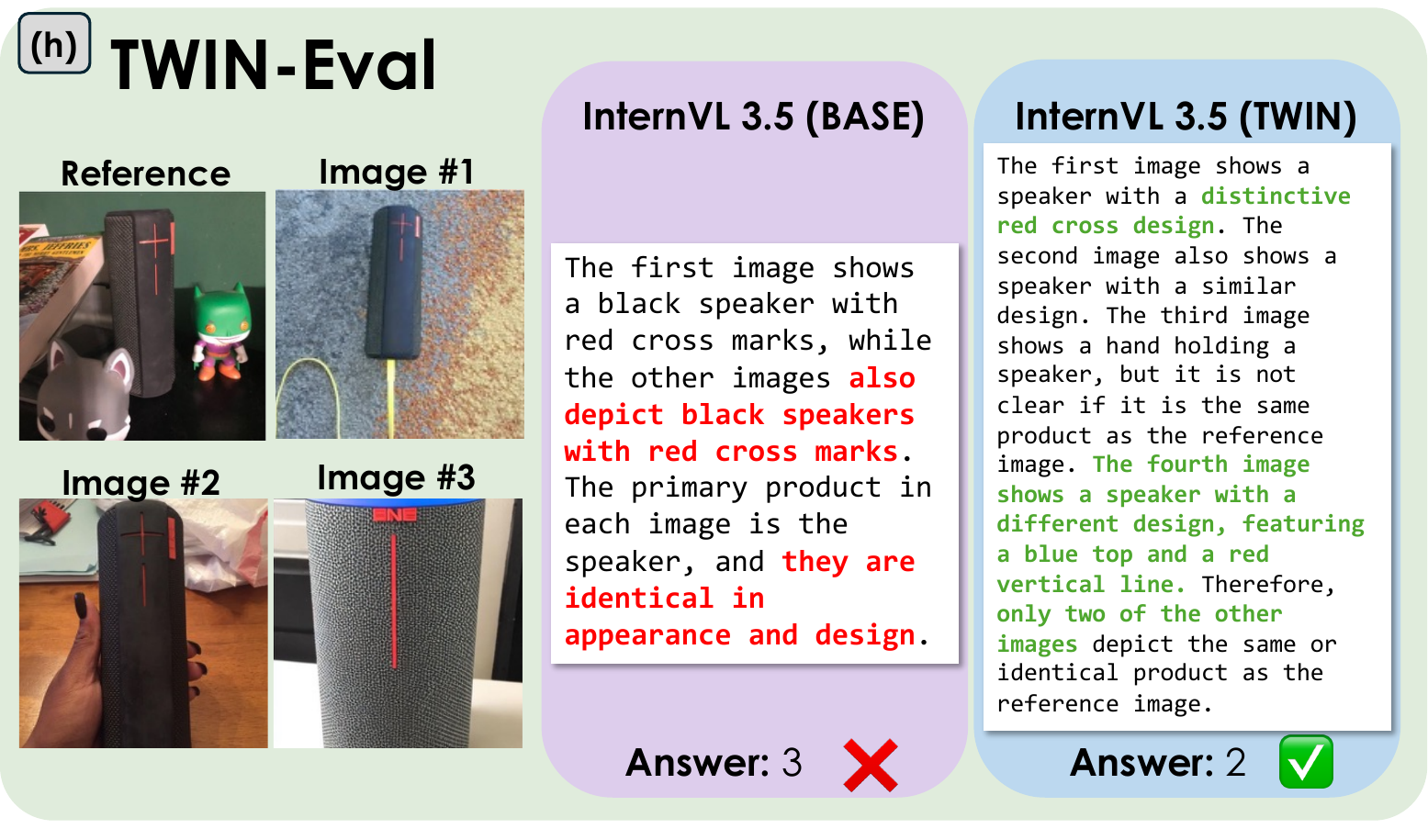}
    \caption{\textbf{Outputs on~\oureval~for base VLMs and variants trained on~\ourtrain.}
    For each example, we show the source dataset, images, and both model predictions. We include both \emph{pair} queries (a-d) and \emph{multi} queries (e-h). Incorrect reasoning is highlighted in \incorrect{red} and correct reasoning in \correct{green}. The \incorrect{red circle} highlights a hard-to-see object. Training on~\ourtrain~enhances fine-grained grounding, leading to increased attention to part geometry and texture over the base model, even for domains that differ from the household objects in~\ourtrain. }
    \label{fig:qualitative_examples}
\end{figure*}

\mypar{Query types.} For \oureval, we construct two query types per dataset. \emph{Pair} queries show two images and ask whether they depict the same instance, artwork, or species. \emph{Multi} queries provide a reference image and three candidates, and ask how many match the reference. Each dataset includes $1,000$ balanced examples per type: pair queries split evenly among positive and negative cases, while multi queries distribute uniformly across answer counts ($250$ each from $0$-$3$). Examples are shown in \cref{fig:fgvqa}, \cref{fig:qualitative_examples}, and the Appendix.

\mypar{Applicability.} \oureval provides a general framework for evaluating fine-grained recognition in VLMs. First, it probes visual perception -- testing whether models can distinguish subtle differences that support accurate reasoning. Second, the pairwise yes/no format aligns with recognition and retrieval tasks, since both 1-of-N classification and retrieval can be cast as a sequence of pairwise comparisons. Together, these properties make \oureval a practical, domain-general suite for analyzing perception in VLMs.
\section{Experiments}
\label{sec:experiments}

We conduct extensive experiments, both quantitative and qualitative, to evaluate the utility of \ourtrain as a training dataset and task for improving fine-grained understanding. Our goals are threefold: (1) establish baseline performance on \oureval, our new fine-grained VQA benchmark suite that repurposes existing datasets, (2) assess if training on \ourtrain improves perception across domains, and (3) ablate the decisions made in our dataset construction and model training.

\mypar{Baselines.} We compare to the following models:

\myparit{(1) Proprietary VLMs}: We establish baseline performance on \oureval for GPT4o~\cite{gpt4} and Gemini 2.5 Flash~\cite{gemini2.5}. 

\myparit{(2) Open-Source VLMs}: We report leading open-source models PerceptionLM-3B~\cite{perceptionlm}, Moondream2-2B~\cite{moondream2}, Gemma3-4B~\cite{gemma3}, and SmolVLM2-2.2B~\cite{smolvlm}. 

\myparit{(3) Direct Comparisons}: We fine-tune Qwen2.5-VL 3B Instruct~\cite{qwen2.5vl} and InternVL3.5 1B Instruct~\cite{internvl3} to evaluate the value of \ourtrain across architectures and scales. Each base model is compared to its trained variant, denoted ``+ \ourtrain''.

\subsection{Main Results}

\cref{table:main_results} shows results on \oureval. Model variants trained on \ourtrain are evaluated zero-shot on all datasets except \ourtrain-Eval. All other models are evaluated zero-shot. All models are prompted without in-context examples. We summarize our findings below.

\mypar{Training on \ourtrain improves fine-grained understanding.} Our direct comparisons show substantial gains from training on \ourtrain. Qwen2.5-VL (3B) achieves strong improvements across datasets: $+18.3\%$ on \ilias~($43.5\%$ to $61.8\%$) and $+17.2\%$ on \ourtrain-Eval~($50.1\%$ to $67.3\%$). InternVL3.5 (1B), despite being smaller and thus less expressive, still shows consistent gains from post-training on \ourtrain: $+3.8\%$ on \cub~($48.3\%$ to $52.1\%$) and $+3.0\%$ on \landmarksshort~($43.8\%$ to $46.8\%$). These consistent improvements across both models indicate that \ourtrain benefits VLMs of different scales and architectures. Importantly, we observe improvement on \emph{multi} queries, which differ in structure from our training data, suggesting improved fine-grained perception rather than task overfitting. 

\mypar{Improvements transfer to unseen domains.} Training on \ourtrain, which features everyday household objects, improves performance even in domains absent from its training distribution. Substantial gains are observed on animal and plant species (\inquire \& \cub), and for art and landmarks (\met \& \landmarksshort), which are distinct from objects in \ourtrain. With Qwen2.5-VL, \inquire accuracy rises by $+19.3\%$ ($54.4\%$ to $73.7\%$), \cub by $+14.4\%$ ($60.7\%$ to $75.1\%$), and~\landmarksshort by $+4\%$ ($53.9\%$ to $57.9\%$).

\mypar{Qualitative Examples.} We present qualitative results in~\cref{fig:qualitative_examples}, comparing baseline models with their counterparts post-trained on \ourtrain. The textual explanations reveal which visual cues models rely on, showing how \ourtrain enhances fine-grained reasoning. In (a), the trained Qwen2.5-VL distinguishes subtle color differences in the numbers and clock hands. In (c), it correctly identifies shared classical columns. In (e), it matches the reference card under challenging lighting in Image~2. In (g), the base model incorrectly flags Image~2 as a different knife, while the trained variant recognizes consistent blade texture and handle material. In (h), only the trained model notices color and design differences in the synthetic negative speaker (Image~3).

These improvements extend to out-of-distribution examples. In (b), the post-trained model identifies the bird by attending to its black head, white body, and orange beak. In (d), it notes differences in date and color scheme. In (f), it detects color differences between the reference bird and Image~3. These improved reasoning descriptions confirm that \ourtrain enhances understanding beyond the training domain.

\mypar{Comparison to open-source VLMs.} Among open-source baselines, Gemma3 performs best across benchmarks: $53.8\%$ on \ourtrain-Eval, $69.4\%$ on \inquire, and $72.7\%$ on~\ilias. However, as Gemma3 does not support training with Flash Attention~\cite{flashattention2,gemma-flash-attn1,gemma-flash-attn2}, fine-tuning it in PyTorch would be prohibitively expensive. Therefore, we train the second-best open-source VLM, Qwen2.5-VL, on \ourtrain. Our trained Qwen2.5-VL model (+ \ourtrain) outperforms Gemma3 on \ourtrain-Eval ($67.3\%$ vs $53.8\%$), \inquire ($73.7\%$ vs $69.4\%$), and \cub ($75.1\%$ vs $73.2\%$). Gemma3 outperforms our trained Qwen2.5-VL model on \met ($72.7\%$ vs $66.0\%$) and \landmarksshort ($73.2\%$ vs $57.9\%$) -- the datasets where we see the least improvement from training.

\mypar{Comparison to proprietary VLMs.} We also benchmark proprietary models on \oureval. GPT4o and Gemini-2.5-Flash outperform all open-source baselines ($81.7\%$ for GPT4o vs $53.8\%$ for Gemma3 on \ourtrain-Eval; $91.4\%$ for Gemini 2.5 Flash vs $67.5\%$ for Gemma3 on \ilias). These models are likely far larger and trained on proprietary data. Training on \ourtrain narrows this gap: Qwen2.5-VL + \ourtrain reaches $73.7\%$ on \inquire, reducing the gap from $29.5\%$ to $10.2\%$, showing that targeted fine-grained training substantially closes the gap with proprietary models.

\subsection{Ablations and Analysis}
\label{subsec:ablations}

\begin{figure}[t!]
    \centering
    \includegraphics[width=0.9\columnwidth]{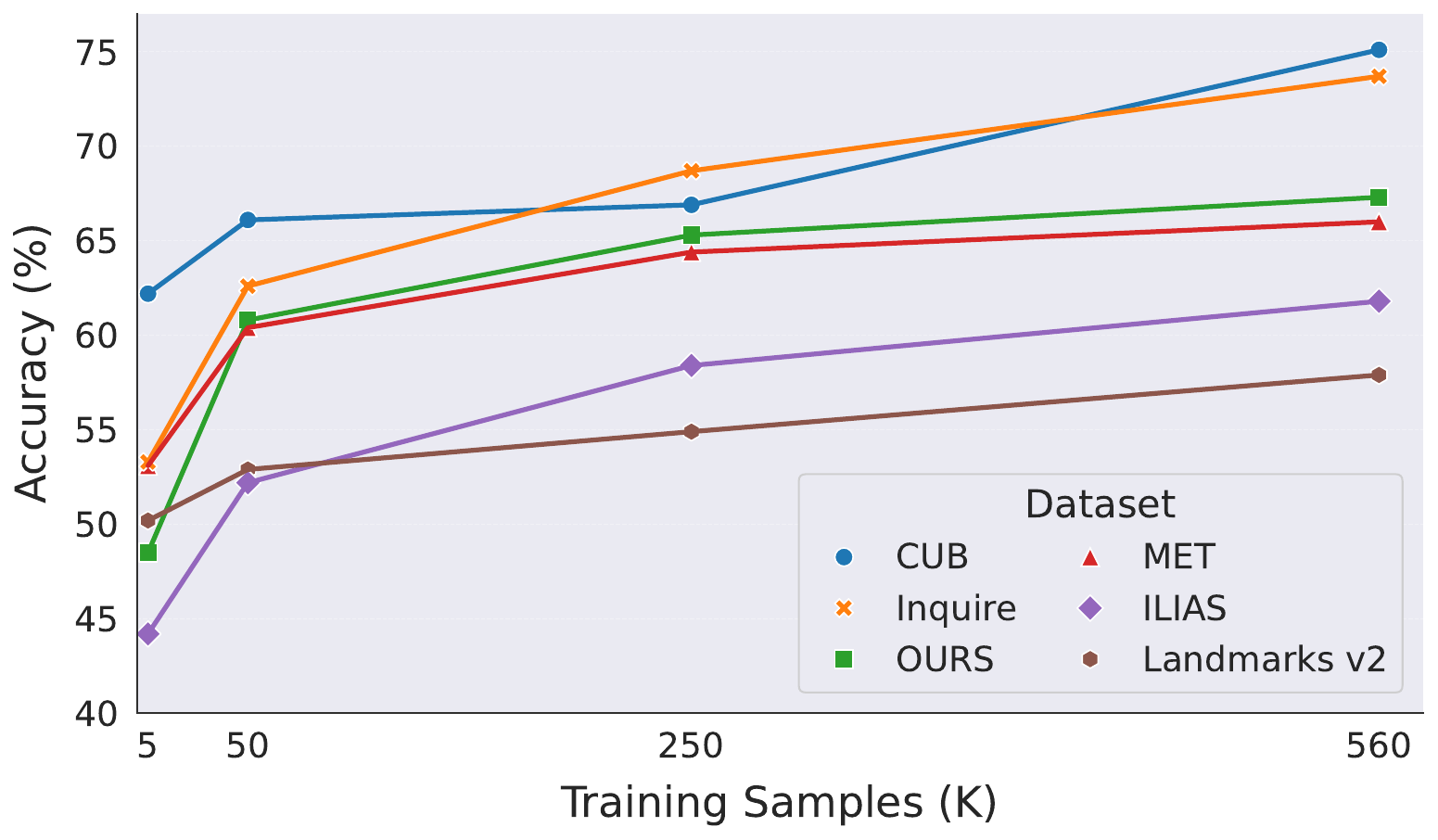}
    \vspace{-3mm}
    \caption{Effect of training set size on~\oureval accuracy (\%).}
    \label{fig:training-scaling}
    \vspace{-3mm}
\end{figure}

We now analyze factors driving improved performance on \oureval. First, we study the \ourtrain dataset and examining choices made during its construction, namely the challenging negative pairs (real and synthetic) in~\cref{table:hard_negatives} and data scale in~\cref{fig:training-scaling}. Next, we compare post-training paradigms on \ourtrain, supervised fine-tuning vs reinforcement learning, in~\cref{table:sft_vs_rl}. Finally, we  probe the vision encoder to showcase that training on \ourtrain improves underlying representations in~\cref{table:embedding_probe}. We summarize our findings below.

\begin{table}[t]
\resizebox{0.99 \linewidth}{!}{
\begin{tabular}{l|c|cccccc}
\toprule
& \textsc{Mean} & \ourtrain-Eval & \ilias & \landmarksshort & \met & \cub & \inquire \\

\midrule
Qwen2.5-VL 3B & 53.0 & 50.1 & 43.5 & 53.9 & 55.5 & 60.7 & 54.4 \\
+ \ourtrain w/o Hard Neg. & 58.6 & 51.1 & 53.1 & \textbf{57.9} & 60.0 & \textbf{68.5} & 60.9\\
+ \ourtrain & \textbf{63.1} & \textbf{65.3 }& \textbf{58.4} & 54.9 & \textbf{64.4} & 66.9 & \textbf{68.7} \\
\bottomrule
\end{tabular}
}
\vspace{-2mm}
\caption{\textbf{Impact of hard negatives.} Models in rows 2 and 3 are trained on the \ourtrain variant of their corresponding row. The hard negatives in \ourtrain significantly improve performance on~\oureval.}
\label{table:hard_negatives}
\vspace{-5mm}
\end{table}
\mypar{Hard Negatives.} We evaluate the contribution of hard negatives to \oureval performance in~\cref{table:hard_negatives}. We post-train Qwen2.5-VL 3B on two variants: (1) \emph{\ourtrain}, a $250$K-sample subset matching the original ratios of positive/negative and real/generated images—our final setting; and (2) \emph{\ourtrain w/o Hard Neg.}, where all hard negatives are replaced with random negatives. This controlled comparison isolates the value of including hard negatives in \ourtrain. We report total accuracy ($\%$) on each \oureval dataset separately, as well as the mean across datasets.

\cref{table:hard_negatives} shows that even without hard negatives, our task yields modest improvements over the base model ($51.1\%$ vs $50.1\%$ on \ourtrain-Eval; $60.0\%$ vs $55.5\%$ on \met). However, incorporating hard negatives improves performance by $4.5\%$ on average ($63.1\%$ vs $58.6\%$). The gains are most pronounced on in-domain datasets like \ourtrain-Eval ($65.3\%$ vs $51.1\%$) and \ilias ($58.4\%$ vs $53.1\%$), but remain substantial on out-of-distribution datasets like \inquire ($68.7\%$ vs $60.9\%$). These results validate our decision to collect human-verified hard negatives, demonstrating they are critical for developing fine-grained visual understanding.

\mypar{Scaling Analysis.}~\cref{fig:training-scaling} shows the impact of data scale for \ourtrain. We train a Qwen2.5-VL-Instruct 3B model with $M = \{5$K$, 50$K$, 250$K$, 561$K$\}$ image pairs and plot accuracy on \oureval. We sample $M$ with the same proportions of positive/negative pairs and real/generated images as our full dataset. Performance improves consistently across all datasets from $5$K to $561$K samples (\eg $48.5\%$ to $67.3\%$ on \ourtrain and $44.2\%$ to $61.8\%$ on \ilias), reinforcing our decision to collect \ourtrain at scale. Notably, scale also improves performance on domains not represented in \ourtrain: \inquire improves from $53.3\%$ to $73.7\%$ and \cub from $62.2\%$ to $75.1\%$ as training data scales from $5$K to $561$K.

\begin{table}[t]
\resizebox{0.99 \linewidth}{!}{
\begin{tabular}{l|c|cccccc}
\toprule
& \textsc{Mean} & \ourtrain-Eval & \ilias & \landmarksshort & \met & \cub & \inquire \\

\midrule
RL & \textbf{57.5} & \textbf{64.0} & 50.3 & \textbf{48.5} & \textbf{58.7} & \textbf{65.0} & \textbf{58.7} \\
SFT & 53.8 & 61.2 & \textbf{53.6} & 47.6 & 52.3 & 53.9 & 54.2\\
\bottomrule
\end{tabular}
}
\vspace{-2mm}
\caption{\textbf{SFT vs. RL.} We tune Qwen2.5-VL 3B on a subset of \ourtrain with SFT and RL. We report total accuracy (\%) on~\oureval.}
\label{table:sft_vs_rl}
\vspace{-4.9mm}
\end{table}

\mypar{SFT vs. RL.} We compare post-training with reinforcement learning (RL) versus supervised fine-tuning (SFT) in~\cref{table:sft_vs_rl}. Our model produces both a chain-of-thought (CoT) explanation and a final yes/no prediction. RL optimizes a reward based solely on the final answer correctness, while SFT performs supervision on all output tokens, including both CoT and final prediction. To obtain high-quality supervision for SFT, we collect responses from Gemini-2.5-Flash~\cite{gemini2.5} on a subset of \ourtrain and retain only those where Gemini is correct. We then post-train Qwen2.5-VL on these $136$K pairs with both SFT and RL. Details for SFT are in the Appendix. Although both methods achieve similar performance on \ourtrain-Eval ($61.2\%$ with SFT vs $64.0\%$ with RL), the RL variant outperforms on out-of-distribution datasets ($53.9\%$ with SFT vs $65.0\%$ with RL on \cub; $52.3\%$ with SFT vs. $58.7\%$ with RL on \met). These results corroborate prior findings that RL better preserves generalization during post-training than SFT~\cite{sftmemorizesrlgeneralizes, rlsrazor, huan2025does, unveilingcompositiongap}.

\begin{table*}[]
\resizebox{0.99 \linewidth}{!}{
\begin{tabular}{lccccccccccc}
\toprule
 & \realworldqa~\cite{realworldqa} & \pope~\cite{pope} & \cvbench~\cite{cambrian1} & \nlvr~\cite{nlvr2} & \aitwod~\cite{ai2d} & \seed~\cite{seedbench} & \chartqa~\cite{chartqa} & \textvqa~\cite{textvqa} & \blink~\cite{blink} & \mmmu~\cite{mmmu} & \vlmbias~\cite{vlmbias} \\

\midrule

\sectionrow{12}{Direct Comparisons}
Qwen2.5 VL 3B~\cite{qwen2.5vl} & 59.8 & 87.6 & 66.0 & 79.8 & 78.6 & 75.0 & 83.4 & \textbf{79.4} & 47.3 & \textbf{47.1} & 17.2 \\
+ \ourtrain & \textbf{61.8} & \textbf{88.2} & \textbf{66.6} & \textbf{81.2} & \textbf{79.4} &  \textbf{75.1}& \textbf{84.1} & 77.9 & \textbf{47.5} & 46.9 & \textbf{17.4} \\

 \midrule

InternVL3.5 1B~\cite{internvl3} & 52.2 & 86.3 & 59.8 & 73.6 & 67.1 & \textbf{70.9} & 63.0 & \textbf{55.9} & 39.3 & \textbf{40.4} & 17.4 \\
+ \ourtrain & \textbf{52.6} & 86.3 & \textbf{60.2} & \textbf{73.8} & \textbf{67.2} & 70.8 & \textbf{63.2} & 55.7 & \textbf{39.7} & 40.1 & 17.4 \\

\bottomrule
\end{tabular}
}
\vspace{-2mm}
\caption{
\textbf{Accuracy (\%) on general VQA benchmarks.} VLMs post-trained on~\ourtrain retain performance on general VQA benchmarks.}
\label{table:general_vqa_results}
\vspace{-5mm}
\end{table*}
\begin{table}[]
\resizebox{0.99 \linewidth}{!}{
\begin{tabular}{l|cc|cc|cc|cc}
\toprule
 & \multicolumn{2}{c|}{\pets~\cite{oxfordpets}} & \multicolumn{2}{c|}{\sun~\cite{sun397}} & \multicolumn{2}{c|}{\cub~\cite{cub}} & \multicolumn{2}{c}{\cifar~\cite{cifar100}}  \\
 & Lin. & KNN & Lin. & KNN & Lin. & KNN & Lin. & KNN  \\

\midrule

Qwen2.5 VL 3B~\cite{qwen2.5vl} & 75.0 & 72.4 & 64.4 & 70.4 & 72.3 & 77.4 & 68.5 & 71.8  \\
+ \ourtrain & \textbf{79.1} & \textbf{74.5} & \textbf{66.4} & \textbf{71.8} & \textbf{72.4} & \textbf{77.7} & \textbf{69.6} & \textbf{73.2} \\
\bottomrule
\end{tabular}
}
\vspace{-2mm}
\caption{\textbf{Encoder probing results.} We evaluate Qwen2.5-VL's vision encoder via linear probing and KNN. Post-training on~\ourtrain yields embeddings better suited for fine-grained classification.}
\vspace{-4mm}
\label{table:embedding_probe}
\end{table}

\mypar{Encoder Probing.} We probe the underlying vision encoder of Qwen2.5-VL to examine whether training on \ourtrain improves its visual representations. Unlike contrastive models (\eg CLIP~\cite{clip}), this encoder was not trained to produce embeddings useful in isolation. Nevertheless, we apply traditional probing methods to assess representation quality and determine if post-training on \ourtrain improves representation ability of vision encoders in VLMs. In~\cref{table:embedding_probe}, we report accuracy on fine-grained classification datasets covering various domains: \pets~\cite{oxfordpets} with cat and dog species, \sun~\cite{sun397} with real world scenes, \cub~\cite{cub} with bird species, and \cifar~\cite{cifar100} with everyday objects and animals. We follow~\cite{perceptionencoder} and evaluate two settings: a linear probe (Lin.) trained on embeddings from each dataset’s training split and then frozen for validation, and K-nearest neighbor retrieval (KNN), where test queries are matched against embeddings from the corresponding training split.

From~\cref{table:embedding_probe}, we find that post-training on~\ourtrain yields consistent improvements across fine-grained classification datasets: $+4.1\%$ with a linear probe on \pets ($75.0\%$ to $79.1\%$), $+1.4\%$ with KNN on \cifar ($71.8\%$ to $73.2\%$), and $+2\%$ with a linear probe on \sun ($64.4\%$ to $66.4\%$). The improvement to probing accuracy across datasets suggest that by forcing VLMs to detect subtle differences in objects, post-training on \ourtrain yields embeddings better suited for fine-grained understanding.

\subsection{Additional Results} 
To ensure post-training on~\ourtrain does not erode general visual reasoning, we report results on popular VQA benchmarks in~\cref{table:general_vqa_results}. These benchmarks span common sense understanding (\seed~\cite{seedbench}, \mmmu~\cite{mmmu}, \nlvr~\cite{nlvr2}), spatial reasoning and grounding (\realworldqa~\cite{realworldqa}, \cvbench~\cite{cambrian1}, \blink~\cite{blink}), hallucination and bias (\pope~\cite{pope}, \vlmbias~\cite{vlmbias}), OCR (\textvqa~\cite{textvqa}), and chart/diagram understanding (\chartqa~\cite{chartqa}, \aitwod~\cite{ai2d}). Although not our main focus, models post-trained on \ourtrain retain performance on general VQA benchmarks, with small improvements on most for both Qwen2.5-VL ($+1.4\%$ on \nlvr; $+0.8\%$ on \aitwod) and InternVL3.5 ($+0.4\%$ on \realworldqa; $+0.4\%$ on \cvbench). 

\begin{table}[t]
\centering
\resizebox{0.99\linewidth}{!}{
\begin{tabular}{lccc}
\toprule
 & \mmlu~\cite{mmlu} & \hellaswag~\cite{hellaswag} & \gsmeightk~\cite{gsm8k} \\

 \midrule

Qwen2.5 VL 3B~\cite{qwen2.5vl} & 64.7 & 66.4 & 70.8 \\
+ \ourtrain & 64.6 & 66.4 & 68.8 \\

\midrule

InternVL3.5 1B~\cite{internvl3} & 49.0 & 48.8 & 57.8 \\
+ \ourtrain & 49.0 & 48.9 & 58.1 \\

\bottomrule
\end{tabular}
}
\vspace{-2mm}
\caption{
\textbf{Accuracy (\%) on text-QA benchmarks.} Post-training on \ourtrain with RL leaves text-QA performance largely unaffected.}
\label{table:text_results}
\vspace{-6mm}
\end{table}

Beyond vision,~\cref{table:text_results} reports performance on popular text-only benchmarks covering different skills: \mmlu~\cite{mmlu} for common knowledge, \hellaswag~\cite{hellaswag} for sentence completion, and \gsmeightk~\cite{gsm8k} for mathematical reasoning. We find that post-training leaves performance largely unchanged on these benchmarks: $64.7\%$ vs $64.6\%$ for Qwen2.5-VL on \mmlu; $57.8\%$ vs $58.1\%$ for InternVL3.5 on \gsmeightk. These results reaffirm prior findings that RL-based tuning preserves model skills~\cite{rlsrazor, sftmemorizesrlgeneralizes}.

\section{Limitations \& Future Work}
\label{sec:limitations}
We introduce \ourtrain, a large-scale VQA dataset of $561{,}000$ queries designed for improving fine-grained perception in VLMs. Each query in \ourtrain asks whether two images depict the same object instance, forcing VLMs to perceive subtle visual cues to answer correctly. \ourtrain includes human-verified hard negative pairs, which we determine are crucial in our experiments. To measure progress on fine-grained understanding, we additionally introduce \oureval, a benchmark suite for precise visual understanding across a wide range of domains. Current open-source VLMs struggle on \oureval, but post-training them on \ourtrain substantially improves fine-grained reasoning, even on unseen domains. Our scaling analysis shows that collecting \ourtrain at scale is key for improved performance. We include a thorough failure analysis of post-trained models in the Appendix.
\noindent There is an extensive list of future work to consider:
\begin{itemize}
    \item We explore GRPO post-training with rewards applied only on final answers. More structured rewards, \eg using multi-modal verifiers, may provide a richer learning signal and lead to improved perception capabilities.
    \item Our experiments show hard negatives in \ourtrain are essential for effective learning. Exploring automated techniques to mine even harder negatives, \eg with the help of model-in-the-loop data engines, is a promising direction.
    \item \ourtrain includes substantial viewpoint variation, requiring models to reason over fine-grained part geometries that may appear different across image pairs. Integrating 3D representations as an explicit means of encoding spatial structure may further improve performance on the task.
\end{itemize}
We envision \ourtrain as a drop-in addition to VLM training corpora and hope \oureval serves as a benchmark suite for measuring progress in fine-grained understanding. \textbf{To support future research in this direction, we will release all data, code, and post-trained models.}
\section*{Acknowledgments}
We thank Aadarsh Sahoo, Ilona Demler, and Ziqi Ma for their feedback on the project. The project is funded by Meta through the LLM evaluation research grant and partly through Caltech’s CAST program. We also thank Google’s Gemma Academic program for granting us API credits for their LLMs.
{
    \small
    \bibliographystyle{ieeenat_fullname}
    \bibliography{main}
}
\clearpage
\setcounter{page}{1}
\maketitlesupplementary

\enablemaintoc
\appendix

\tableofcontents

\begin{figure*}[t]
    \centering
    \includegraphics[width=0.95\linewidth]{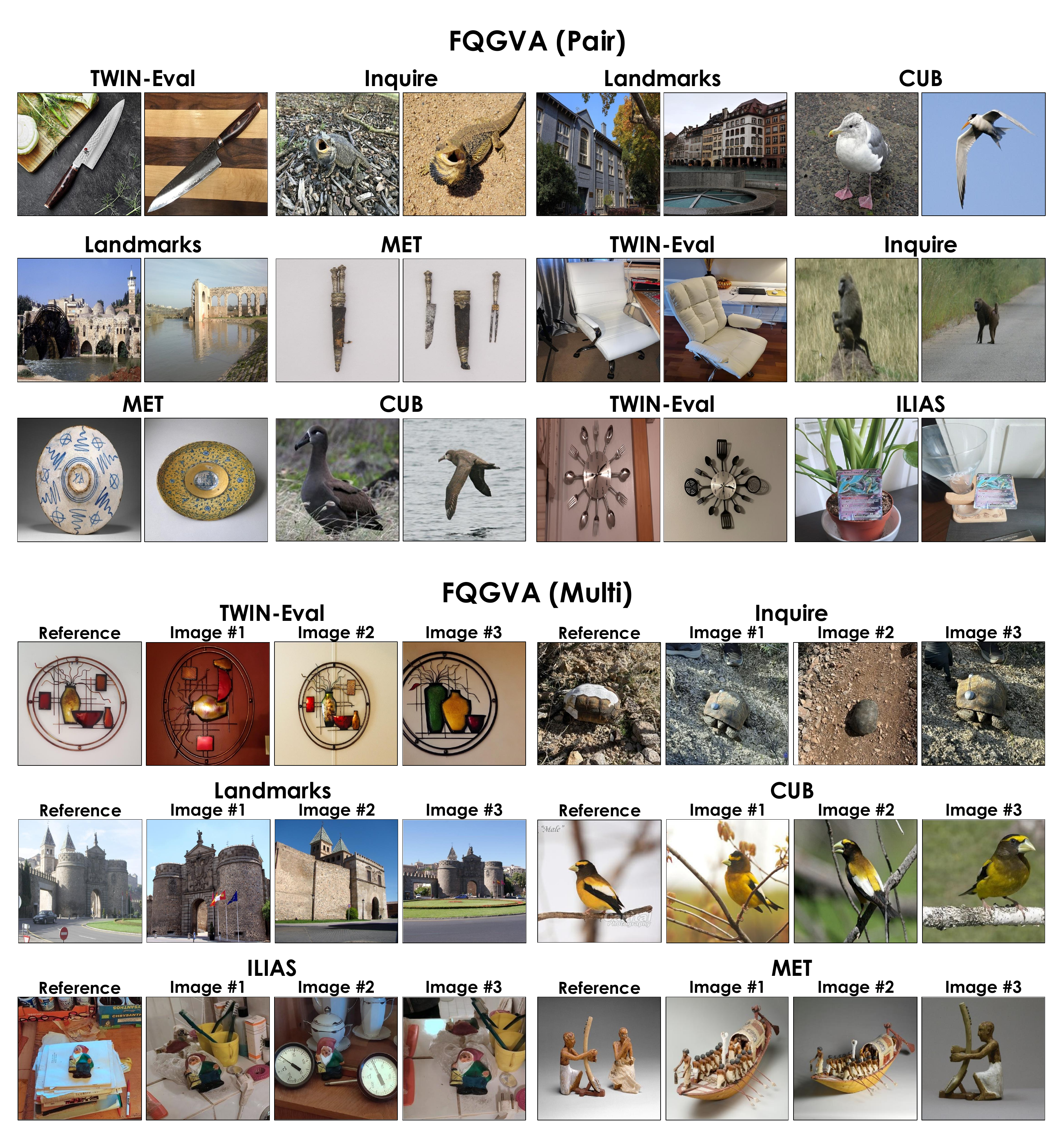}
    \vspace{-3mm}
    \caption{\textbf{Additional examples from \oureval.} We show examples of images in \emph{pair} and \emph{multi} queries. \oureval spans six datasets across a wide range of domains, creating a challenging suite for assessing cross-domain fine-grained VQA in VLMs.}
    \vspace{-4mm}
    \label{fig:appx_fgqva_examples}
\end{figure*}

\section{Additional details on~\oureval}
\label{appx:fgvqa}
We provide additional details on \oureval, our new benchmark suite for evaluating fine-grained VQA. We describe the datasets included in \oureval in~\cref{appx:fgvqa_bmarks}. We illustrate additional examples in~\cref{fig:appx_fgqva_examples}, and describe all prompts used in~\cref{appx:fgvqa_prompts}. We report exact match accuracy for all benchmarks on both \emph{pair} and \emph{multi} queries.

\subsection{Evaluation Benchmarks}
\label{appx:fgvqa_bmarks}
We describe the provenance of each benchmark in \oureval below.

\mypar{\ourtrain-Eval} is the evaluation set of \ourtrain. It is collected identically to \ourtrain (\cref{sec:dataset}), but features distinct instances and images from \ourtrain.

\mypar{ILIAS}~\cite{ilias} is a large-scale test dataset of instance-level image retrieval. It predominantly features images of retail products taken in various contexts, backgrounds, and lighting. For both \emph{pair} and \emph{multi} queries, we source images from the \texttt{core\_db} split, using the \texttt{\_\_key\_\_} field to determine instance identity.

\mypar{Google Landmarks v2 (\landmarksshort)}~\cite{landmarksv2} is a landmark recognition dataset featuring human-made and natural landmarks. The dataset has been used in both classification and retrieval settings. To ensure sufficient images per landmark to for \emph{multi} queries, we source images from the \texttt{train} split and use the \texttt{label} field to determine landmark identity.

\mypar{MET}~\cite{met} is an image retrieval dataset featuring artwork from the Metropolitan Museum of Art in New York. The dataset features images of the same art piece or sculpture from varying viewpoints, emphasizing multi-view consistency in retrieval. We use the \texttt{mini\_met} set to construct both \emph{pair} and \emph{multi} queries.

\mypar{CUB}~\cite{cub} is a fine-grained classification dataset that focuses on identifying bird species from images. We use the \texttt{test} split for all queries and create \emph{pair} and \emph{multi} queries based on the specie of bird.

\mypar{Inquire}~\cite{inquire} is a benchmark for natural world image retrieval, featuring images of animal and plant species sourced from iNaturalist~\cite{inaturalist}. We source images from the \texttt{validation} split and use the \texttt{inat24\_species\_name} field to determine instance identity for both \emph{pair} and \emph{multi} queries.

\subsection{\textbf{\oureval} Prompts}
\label{appx:fgvqa_prompts}
We include all prompts used on all datasets for both \emph{pair} and \emph{multi} queries in~\crefrange{fig:prompts_twin_eval}{fig:prompts_inquire}. All prompts are identical in format, except for the domain-specific instance names (\eg bird/object/artwork/species). Additionally, for benchmarks featuring object instances (\ourtrain-Eval, \ilias~\cite{ilias}), we found it necessary to provide the definition of an object instance as models were incorrectly labeling different objects of the same-category (\eg two earbuds of different colors) as the same.

\begin{table}[t]
\resizebox{0.99 \linewidth}{!}{
\begin{tabular}{l|c|cccccc}
\toprule
& \textsc{Mean} & \ourtrain-Eval & \ilias & \landmarksshort & \met & \cub & \inquire \\

\midrule
Human & 92.9 & 90.0 & 100.0 & 82.5 & 93.8 & 92.5 & 98.8 \\
\bottomrule
\end{tabular}
}
\vspace{-2mm}
\caption{\textbf{Human Evaluations on \oureval}.}
\label{table:human_evals}
\vspace{-5mm}
\end{table}
\subsection{\textbf{\oureval} Human Evaluation}
\label{appx:human_evals}
We establish human performance on \oureval in~\cref{table:human_evals}. We task four human annotators with repsonding to a subset of 20 queries per benchmark in \oureval. We report total accuracy (\%) on each dataset, averaged across evaluator. We find that human evaluators significantly outperform open-source VLMs, suggesting ample future work is needed in fine-grained VQA.

\section{Additional details on~\ourtrain}
\label{appx:twin}
We provide further details on the collection of \ourtrain, including instance sourcing, de-duplication, hard negative pair assignment, and details on generating synthetic negatives.

\subsection{Instance Sourcing}
We used the Amazon Reviews dataset \cite{amazon-reviews} to extract a large set of product listings, which would serve as candidate objects, and associated noisy review images. We categorized these objects into supercategories based on their associated category in the Amazon store (\eg ``Electronics") and further into categories based on the product title (\eg ``Speaker"). We aimed to maximize diversity in our instances and thus selected 36 categories to draw from. We subsequently tasked human annotators with removing products that either lacked visual consistency (\eg the same product being sold in multiple colors) or did not adhere to the object category (\eg ``earbud replacement tips" instead of ``earbuds").

\begin{figure*}[t!]
    \centering
    \includegraphics[width=0.9\linewidth]{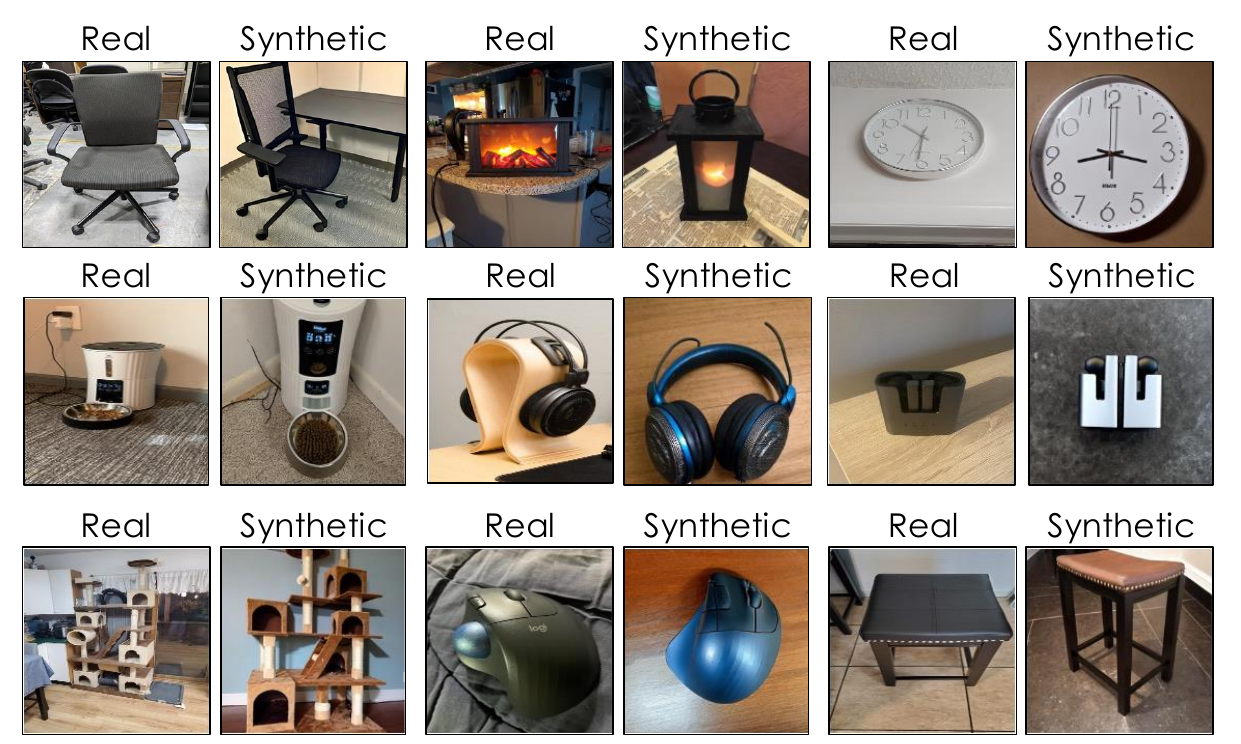}
    \vspace{-3mm}
    \caption{\textbf{Additional Synthetic Negatives.} For each pair, we show a real image of the object instance and a synthetic image of the same instance generated with Dreambooth~\cite{dreambooth}. We find the synthetic negatives capture the ``gist" of the instance, but are not faithful to fine details such as color, geometry, and texture -- making them a strong additional set of hard negative pairs.}
    \vspace{-4mm}
    \label{fig:appx_gen_neg}
\end{figure*}

\subsection{De-Duplication} 
To ensure the correctness of the pairwise labels in \ourtrain, we must ensure that object instances are \emph{unique}. If object instances are not unique, then two of the same objects could incorrectly be listed as negative pairs. However, on Amazon duplicate listings are common. We therefore merged duplicate objects into the same instance. To this end, we used a pre-trained CLIP model to compare the images of each product with the images of all other products via cosine similarity. We flagged the $k$ images with the highest similarity as potential duplicates, setting $k$ equal to the number of images for the product. We then tasked human annotators with manually inspecting the flagged images. If a duplicate was found, we merged the two products. The resulting object instances after de-duplication are used to produce the $561$K instances of \ourtrain.

\subsection{Hard Negative Pair Assignment}
Once we have collected a set of object instances, we construct hard negative pairs -- pairs of distinct objects that appear similar. We collect these hard negatives with the help of human annotators. First, using CLIP~\cite{clip}, for each object instance, we compute pairwise cosine similarities between the image embeddings of that instance and all other instances. We use these similarities to shortlist $k$ visually similar candidates for each instance, with $k$ being a random number between $1-2\times$ the number of positive images of that instance. Human annotators then select the final pairs, omitting pairs they deem to be too easy.

\subsection{Additional Details on Generated Negatives}
We augment the set of negative pairs with additional synthetically generated images. Augmenting with generated negatives allows us to produce additional negative pairs without extra data collection. We follow the procedure outlined in Dreambooth~\cite{dreambooth}. For each object instance, we sample 3 to 5 images for each instance to serve as grounding images. We set the initializer tokens to the generic category associated with the instance (\eg ``vase", ``headphones"). Our prompts to Dreambooth are: [``an image of \{category\}", ``similar to \{category\}", ``a picture of\{category\}, ``show me a \{category\}, ``here is a \{category\}"] replacing \{category\} with the category of the instance. Finally, we task human annotators with validating the generated samples. Annotators are asked to remove generations that either do not appear similar to the instance or are indistinguishable from the instance -- yielding a set of hard negative pairs. 

We show additional examples of synthetic negative pairs in~\cref{fig:appx_gen_neg}. For each pair of images, we show the real image on the left and the paired synthetic image on the right. We find that synthetic negatives faithfully capture the ``gist" and overall appearance of the object instance, but they fail to capture nuanced details such as part color and geometry, texture, and brand text/logos. These subtle differences in appearances yield an augmenting set of hard negative pairs.

\section{Additional Training Details}
\label{appx:training}
We include additional details for all model training. In~\cref{appx:grpo}, we provide an overview of GRPO~\cite{deepseekmath}, our main post-training method, and our training setup. We provide additional details on supervised fine-tuning (\cref{subsec:ablations}) in~\cref{appx:sft}. We include all hyperparameters in~\cref{appx:grpo_params} and \cref{appx:sft_params}.

\subsection{GRPO}
\label{appx:grpo}
To post-train VLMs on \ourtrain, we use Group Relative Policy Optimization (GRPO;~\cite{deepseekmath}). We provide an overview of this method below.

\mypar{Setup.} Given a pair of images $(I_1, I_2)$ with ground-truth pairwise label $y \in \{\text{yes}, \text{no}\}$, our base VLM $\pi_\theta$, parametrized by $\theta$, is prompted to produce a reasoning explanation $r$ and final answer $\hat{y}$ whether both images depict the same instance. We wish to find $\pi_{\theta^*}$ that correctly identifies pairs of images that show the same instance. We include the VQA prompt used during training in~\cref{fig:prompts_training}.

\mypar{Reward Design.} We define a simple binary outcome reward $R$ that compares the predicted final answer with the ground truth pairwise assignment: $R(y, \hat{y}) = \mathbf{1}_{\{y = \hat{y}\}}$. Our supervision thus relies \emph{only} on pairwise assignments and does not use any descriptive textual annotations. 

\begin{figure*}[t!!]
    \centering
    \includegraphics[width=0.98\linewidth]{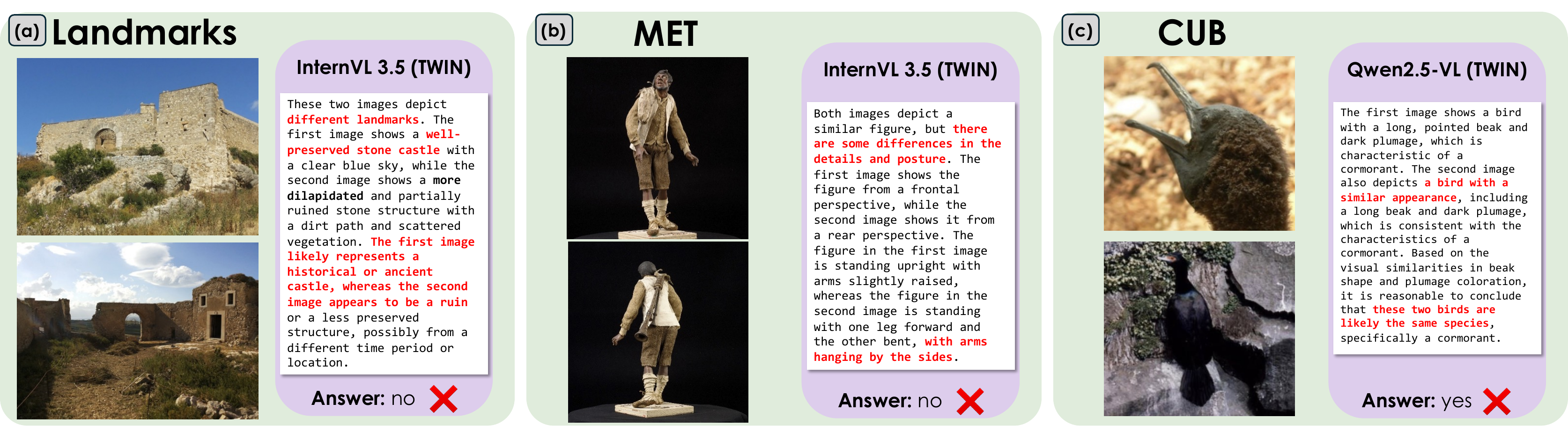}
    \includegraphics[width=0.98\linewidth]{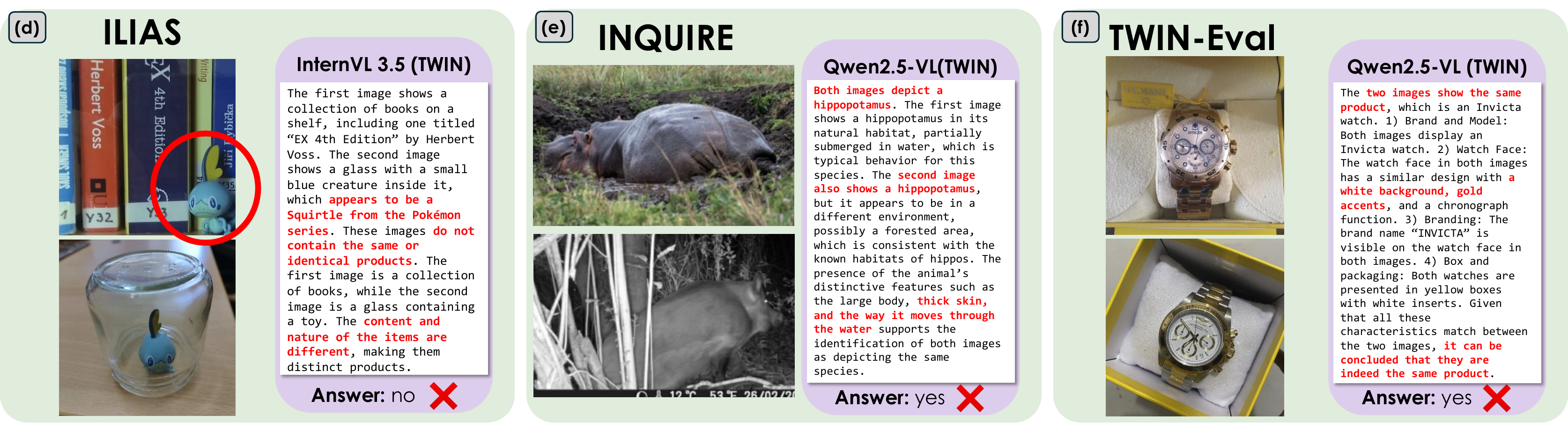}
    \caption{\textbf{Failure Cases on~\oureval~for models post-trained on~\ourtrain.}
    For each example, we show the source dataset, images, and model predictions. Incorrect reasoning is highlighted in \incorrect{red}. The \incorrect{red circle} highlights hard-to-see objects.}
    \label{fig:appx_failures}
\end{figure*}

\mypar{Optimization.} We optimize our VLM $\pi_\theta$ via GRPO~\cite{deepseekmath}. For each image pair $(I_1, I_2)$ with ground truth label $y$, we sample $G$ responses: $\mathcal{O} = \{( I_1, I_2, r^{(i)}, \hat{y}^{(i)})\}^G_{i=1}$ and compute centralized advantages:

\begin{equation}
    A^{(i)} = R(y, \hat{y}^{(i)}) - \frac{1}{G} \sum_{j=1}^G R(y, \hat{y}^{(j)})
\end{equation}

The GRPO objective then directly maximizes expected advantages while maintaining policy stability:

\begin{equation}
\begin{aligned}
\mathcal{L}(\theta)
  &= \frac{1}{G} \sum_{i=1}^G
     \Big(
       \hat{A}^{(i)}
       - \beta\,
         \text{KL}\big[
           \pi_{\theta}
           \,\big\|\,
           \pi_{\text{ref}}
         \big]
     \Big) \\
\hat{A}^{(i)}
  &= \min\Big(
       s^{(i)} A^{(i)},\,
       \text{clip}\big(
         s^{(i)},\,
         1 - \epsilon,\,
         1 + \epsilon
       \big)
       A^{(i)}
     \Big) \\
\end{aligned}
\end{equation}
     
\noindent where $s^{(i)} = \frac{\pi_{\theta}(r^{(i)}, \hat{y}^{(i)} | I_1, I_2)}{\pi_{\text{old}}(r^{(i)}, \hat{y}^{(i)} | I_1, I_2)}$ represents the importance ratio, $\epsilon = 0.2$ is the clipping parameter, and $\beta = 0.01$ is the KL penalty coefficient. The formulation encourages improved fine-grained understanding while simultaneously preventing drift from the pre-trained base policy $\pi_{\text{ref}}$.

\mypar{Implementation.} We train Qwen2.5-VL-3B-Instruct~\cite{qwen2.5vl} and InternVL3.5-1B-Instruct~\cite{internvl3} on \ourtrain, chosen as leading open-source VLMs at the 3B and 1B scales. We do not freeze any part of the model, and train on $4$ A100 GPUs for 1 epoch. We use a batch size of $480$, group size $5$ and learning rate $10^{-6}$. We build on the verl~\cite{verl} repository for training. We detail all hyperparameters in~\cref{appx:grpo_params}. Post-training the Qwen2.5-VL 3B model for one epoch on the $560$K samples of TWIN took approximately 140 hours on 4 A100 GPUs.

\subsection{Supervised Fine-Tuning}
\label{appx:sft}
We compare post-training methods in~\cref{subsec:ablations}. We provide additional details for supervised fine-tuning (SFT) used in those experiments. As our model produces both an explanation and final prediction, SFT requires supervision beyond the pairwise assignment labels used in RL. To obtain high-quality supervision, we prompt Gemini-2.5-Flash~\cite{gemini2.5} with samples from \ourtrain and retain responses where Gemini is correct. We use default generation parameters for Gemini and collect a total of $136$K high-quality answers.

We train Qwen2.5-VL 3B end-to-end using SFT on the collected samples. We use LLaMa-Factory~\cite{llamafactory} for training, using a learning rate of $10^{-6}$ with a cosine scheduler. We train the SFT model for $2$ epochs, as this yielded better accuracy on \oureval. All hyperparameters are in~\cref{appx:sft_params}.

\begin{figure*}[t]
    \centering
    \includegraphics[width=0.8\linewidth]{figures/btw.pdf}
    \vspace{-2mm}
    \caption{Examples from Birds-to-Words~\cite{birdstowords}. The dataset features $3.3$K pairs of \emph{different} bird species, along with a human-annotated description of their differences. The differences are often coarse-grained (\eg Pair 3, Pair 6 -- which both display significant color differences), and the text descriptions subjective (\eg Pair 4 -- where animal 1 is described as "more beautiful" than animal 2).}
    \vspace{-1mm}
    \label{fig:appx_btw_examples}
\end{figure*}

\section{Failure Cases}
\label{appx:failure}
We include failure cases for models post-trained with \ourtrain in~\cref{fig:appx_failures}. For each example, we show the source dataset, images, and model predictions. We find that although models post-trained with \ourtrain demonstrate improved fine-grained understanding, they struggle with fine differences in color (\eg in (a) where the ruins appears different colors due to lighting, and (f) where the watch accents differ between blue and gold). Moreover, extreme viewpoint variation, as seen in (b) remains a challenge. Lastly, we find that models struggle to extrapolate from incomplete views of the animals in (c) and (e). These challenging examples additionally highlight the difficulty of our new \oureval benchmark suite.

\mypar{InternVL on Inquire.} We observe a slight decrease in performance on \inquire ($-1.2\%$) when post-training InternVL3.5 1B on \ourtrain. We investigate the model responses and find that post-training the 1B model on \ourtrain reduces ``direct identification" (\eg, naming a species) and emphasizes comparisons of visual cues (\eg, the color of the feathers). While this improves instance-level reasoning, it can fail under challenging lighting or occlusion, where the base model succeeds by relying more on priors.

\section{Additional Evaluations}
\label{appx:addn_evals}
We provide additional evaluations for models post-trained with \ourtrain. In~\cref{appx:addn_bmarks} we explore the impact of post-training on \ourtrain on the related tasks of monument doppelganger detection and evaluate sensitivity to color and shapes. In~\cref{appx:icl} we compare post-training to prompting with in-context-learning examples.

\subsection{Additional Benchmarks}
\label{appx:addn_bmarks}
We evaluate on an additional set of benchmarks to determine if post-training on TWIN improves robustness to other image variations. We compare the baseline Qwen2.5-VL 3B model with our \ourtrain post-trained variant on monument doppelganger detection~\cite{cai2023doppelgangers}, the "yes or no" queries on shape/color sensitivity from VLMs Eye~\cite{nam2025vlm}, and $500$ pairwise queries from CUTE~\cite{kotar2023these} in~\cref{table:addn_bmarks}.

\begin{table}[h!]
\centering
\vspace{-2mm}
\resizebox{0.99\columnwidth}{!}{
\begin{tabular}{c|cccc}
& Doppel.~\cite{cai2023doppelgangers} & Shape Sens.~\cite{nam2025vlm} & Color Sens.~\cite{nam2025vlm} & CUTE~\cite{kotar2023these} \\ \hline
Qwen2.5-VL 3B & 50.8 & 56.8  & 96.1 & 58.8  \\
+ \ourtrain & \textbf{55.2} & \textbf{95.7} & \textbf{99.7} & \textbf{68.0}
\end{tabular}
}
\vspace{-2mm}
\caption{\textbf{Additional benchmarks.}}
\vspace{-3mm}
\label{table:addn_bmarks}
\end{table}

\noindent Post-training on \ourtrain improves doppelganger detection ($+4.4\%$), despite \ourtrain not including monuments. This mirrors the gains on \landmarksshort and \met seen in~\cref{table:main_results}. Similarly, post-trained models improve at detecting dissimilar shapes ($+38.9\%$), reach near-perfect performance at comparing colors ($99.7\%$) and are more robust to photometric variations ($+9.2\%$ on CUTE). These results reaffirm that \ourtrain pushes models to attend to subtle visual cues over coarse semantic similarity.

\subsection{In-Context Learning}
\label{appx:icl}
We report results on \oureval for in-context-learning (+ICL) prompting in~\cref{tab:icl} to explore if we can achieve improved performance via sophisticated prompting as opposed to post-training. We randomly sample one positive and one negative pair from \ourtrain as in-context examples for each \oureval query.

\begin{table}[h!]
\centering
\vspace{-2mm}
\resizebox{0.99\columnwidth}{!}{
\begin{tabular}{c|c|cccccc}
& Mean & TWIN-Eval & ILIAS & Landmarks & MET & CUB & Inquire \\ \hline
Qwen2.5-VL 3B (+ ICL) & 39.5 & 39.0 & 37.3 & 37.5 & 38.5 & 43.5 & 41.2 \\
Qwen2.5-VL 3B (+ TWIN) & \textbf{67.0} & \textbf{67.3} & \textbf{61.8} & \textbf{57.9} & \textbf{66.0} & \textbf{75.1} & \textbf{73.7} 
\end{tabular}
}
\vspace{-2mm}
\label{tab:icl}
\caption{\textbf{In-Context Learning.}}
\end{table}

\noindent Post-training significantly outperforms in-context-learning ($67.0\%$ vs $39.5\%$ on average). In the ICL setting, the 3B model appears to infrequently misunderstand the order of images in the prompt, or ignores the examples entirely.

\begin{figure*}[t]
    \centering
    \includegraphics[width=0.8\linewidth]{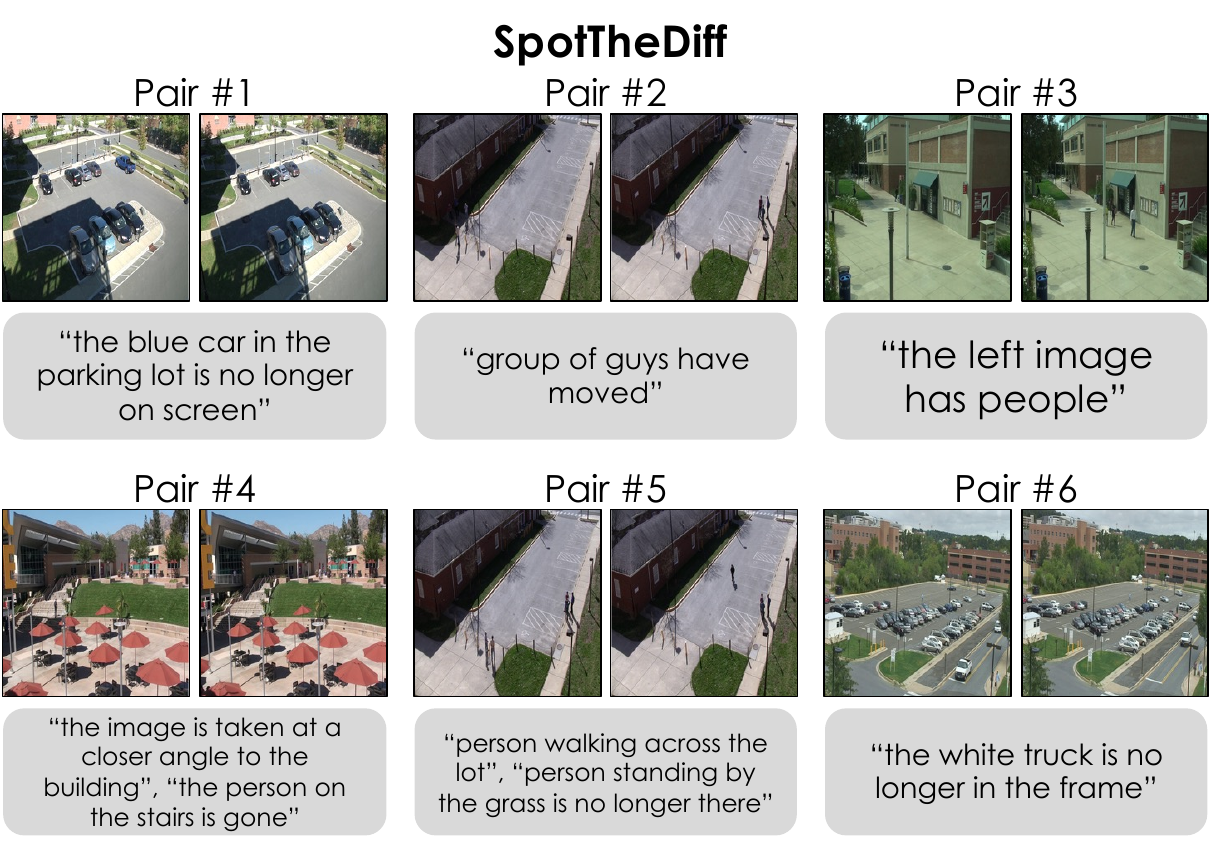}
    \vspace{-2mm}
    \caption{Examples from SpotTheDiff~\cite{spotthediff}. The dataset features $13$K pairs of images taken from different frames of the same video. Unlike \ourtrain, that emphasizes fine differences in object texture and geometry, the differences in SpotTheDiff are primarily spatio-temporal (\eg Pair 1 with the blue car leaving the frame, Pair 2 with the people moving) and highly coarse-grained (\eg Pair 3).}
    \vspace{-1mm}
    \label{fig:appx_std_examples}
\end{figure*}

\section{Comparisons to Prior Datasets}
\label{appx:dataset_comparisons}
We provide a more detailed comparison to prior datasets Birds-To-Words~\cite{birdstowords} and SpotTheDiff~\cite{spotthediff}. While both of these datasets also feature pairs of images as input (similar to \ourtrain), there are several key differences. First, the task in these datasets is to identify the differences between two images, with the assumption \emph{a priori} that the images are indeed different. In contrast, \ourtrain includes both negative pairs and positive pairs in which the instance is unchanged, requiring models to reason about both similarities and differences rather than only locating discrepancies. Second, the differences captured in \ourtrain are substantially more fine-grained, often involving subtle variations in texture or part geometry that go beyond the relatively coarse differences in prior datasets. Third, our dataset is considerably larger in scale ($561$K pairs vs.\ $13$K for SpotTheDiff and $3.3$K for Birds-To-Words). We provide examples and describe the differences with each individual dataset in more detail below.

\subsection{Birds-To-Words}
Birds-to-words~\cite{birdstowords} features $3.3$K pairs of images of different bird species, along with human-annotated descriptions of the differences. We show examples from the dataset in~\cref{fig:appx_btw_examples}. We observe that, unlike the fine-grained differences in \ourtrain, the differences are often much coarser (\eg Pair 3 and Pair 6, which both feature significant color differences). Moreover, the human-annotated descriptions can be subjective (\eg Pair 4 which describes one bird as ``more beautiful").

Compared to \ourtrain, the dataset is orders of magnitude smaller ($3.3$K pairs vs the $561$K in \ourtrain) and spans only bird species, while \ourtrain features a wide range of object categories with significantly different geometries (\eg earbuds and mugs). Additionally, Birds-to-Words features almost exclusively \emph{negative} pairs. While a small number of pairs are annotated as depicting the same instance, these are anomalous cases and not a core aspect of the dataset -- the task is fundamentally to describe the differences between two images. In contrast, \ourtrain emphasizes both positive and negative pairs, requiring models to recognize not only differences but also similarities.

\subsection{Comparison to SpotTheDiff}
SpotTheDiff~\cite{spotthediff} comprises $13$K pairs of images sourced from two different frames of the same video. The videos used depict large scenes from security camera footage. As a result, the differences between frames are predominantly spatio-temporal, concerning which objects or people are visible or no longer visible in the scene (\eg a car leaving the frame, a person moving positions). We visualize examples from SpotTheDiff~\cite{spotthediff} in~\cref{fig:appx_std_examples}. These examples highlight that differences are generally coarse-grained and, unlike the examples in \ourtrain, do not require reasoning about viewpoint variation, part-level geometry, or fine-grained texture changes. Similar to Birds-to-Words, SpotTheDiff assumes the paired images are different, whereas \ourtrain additionally emphasizes recognizing similarities through the inclusion of positive pairs. Finally, \ourtrain is orders of magnitude larger in scale (561K vs.\ 13K pairs) and introduces challenges not present in SpotTheDiff, including substantial viewpoint variation, lighting changes, and differences in context and background.

\clearpage

\newtcblisting{psmall}{
  enhanced jigsaw,
  breakable,
  listing only,
  blank,
  borderline={1pt}{-5pt},
  listing options={breakindent=0pt, breaklines=true, basicstyle=\ttfamily\tiny}
}
\begin{figure*}[t]
\centering
\begin{minipage}{\linewidth}
\begin{psmall}
Pair: Are these images of the same or identical products? For two products to be considered identical, minor changes such as those that can be explained context, backgrounds or photography conditions are allowed, but characteristic features of the product (color, shape, size, etc.) should remain consistent. For images with multiple products, compare only the primary product. 
Explain your reasoning and then conclude with a yes or no answer in <answer> tags as <answer>yes</answer> or <answer>no</answer>.

Multi: The first image is a reference image. How many of the other images depict the same or identical products as the reference image? For two products to be considered identical, minor changes such as those that can be explained by context, backgrounds or photography conditions are allowed, but characteristic features of the product (color, shape, size, etc.) should remain consistent. For images with multiple products, compare only the primary product. 
Explain your reasoning and then answer with a number from 0 to 3 in <answer> tags as <answer>n</answer>.
\end{psmall}
\end{minipage}
\caption{\textbf{\oureval Prompts:} \ourtrain-Eval.}
\label{fig:prompts_twin_eval}
\end{figure*}
\begin{figure*}[t] 
\centering
\begin{minipage}{\linewidth}
\begin{psmall}
Pair: Do these images contain the same or identical products? For two products to be considered identical, minor changes such as those that can be explained by context, backgrounds or photography conditions are allowed, but characteristic features of the product (color, shape, size, etc.) should remain consistent. 
Explain your reasoning and then answer with a yes or no answer in <answer> tags as <answer>yes</answer> or <answer>no</answer>.

Multi: The first image is a reference image. How many of the other images contain the same or identical products to one in the reference image? For two products to be considered identical, minor changes such as those that can be explained by context, backgrounds or photography conditions are allowed, but characteristic features of the product (color, shape, size, etc.) should remain consistent. 
Explain your reasoning and then answer with a number from 0 to 3 in <answer> tags.
\end{psmall}
\end{minipage}
\caption{\textbf{\oureval Prompts:} \ilias~\cite{ilias}.}
\label{fig:prompts_ilias}
\end{figure*}
\begin{figure*}[t]
\centering
\begin{minipage}{\linewidth}
\begin{psmall}
Pair: Do these two images contain the same landmark? 
Explain your reasoning then answer with a yes or no answer in <answer> tags as <answer>yes</answer> or <answer>no</answer>.

Multi: The first image is a reference image. How many of the other images contain the same landmark as the reference image?
Explain your reasoning then answer with a number from 0 to 3 in <answer> tags.
\end{psmall}
\end{minipage}
\caption{\textbf{\oureval Prompts:} \landmarksshort~\cite{landmarksv2}.}
\label{fig:prompts_landmarks}
\end{figure*}
\begin{figure*}[t]
\centering
\begin{minipage}{\linewidth}
\begin{psmall}
Pair: Do these two images contain the same piece of art? 
Explain your reasoning and then answer with a yes or no answer in <answer> tags as <answer>yes</answer> or <answer>no</answer>.

Multi: The first image is a reference image. How many of the other images contain the same piece of art as the reference image?
Explain your reasoning and then answer with a number from 0 to 3 in <answer> tags.
\end{psmall}
\end{minipage}
\caption{\textbf{\oureval Prompts:} \met~\cite{met}.}
\label{fig:prompts_met}
\end{figure*}
\begin{figure*}[t] 
\centering
\begin{minipage}{\linewidth}
\begin{psmall}
Pair: Do these two images show a bird of the same species? 
Explain your reasoning and then conclude with a yes or no answer in <answer> tags as <answer>yes</answer> or <answer>no</answer>.

Multi: The first image is a reference image. How many of the other images show a bird of the same species as the reference image? 
Explain your reasoning and then conclude with a number from 0 to 3 in <answer> tags.
\end{psmall}
\end{minipage}
\caption{\textbf{\oureval Prompts:} \cub~\cite{cub}.}
\label{fig:prompts_cub}
\end{figure*}
\begin{figure*}[t]
\centering
\begin{minipage}{\linewidth}
\begin{psmall}
Pair: Do these two images show an animal or plant of the same scientific species? 
Explain your reasoning and then answer with a yes or no answer in <answer> tags as <answer>yes</answer> or <answer>no</answer>.

Multi: The first image is a reference image. How many of the other images show an animal or plant of the same scientific species as the reference image? 
Explain your reasoning and then answer with a number from 0 to 3 in <answer> tags.
\end{psmall}
\end{minipage}
\caption{\textbf{\oureval Prompts:} \inquire~\cite{inquire}.}
\label{fig:prompts_inquire}
\end{figure*}
\begin{figure*}[t]
\centering
\begin{minipage}{\linewidth}
\begin{psmall}
Are these images of the same or identical products? For two products to be considered identical, minor changes such as those that can be explained by context, backgrounds or photography conditions are allowed, but characteristic features of the product (color, shape, size, etc.) should remain consistent. For images with multiple products, compare only the primary product. Explain your reasoning and then conclude with a yes or no answer in <answer> tags as <answer>yes</answer> or <answer>no</answer>.
\end{psmall}
\end{minipage}
\caption{\textbf{Prompt used for training with \ourtrain.}}
\label{fig:prompts_training}
\end{figure*}

\begin{table*}[t]
  \centering
  \begin{minipage}[t]{0.35\linewidth}
  \centering
  \small
    \begin{tabular}{@{} l r @{}} 
    \toprule
    \textbf{Hyperparameter} & \textbf{Value} \\
    \midrule
    Batch size                            & 480 \\
    Group size                            & 5  \\
    Max prompt length                     & 2048 \\
    Max response length                   & 2048 \\
    Learning rate                         & 1e-6 \\
    Optimizer                             & AdamW \\
    Weight decay                          & 0.01 \\
    KL coefficient                        & 0.01 \\
    Clip ratio                            & 0.2 \\
    Epochs                                & 1 \\
    Rollout engine                        & vLLM \\
    Temperature                           & 1.0 \\
    Top-p                                 & 1.0 \\
    Gradient clipping                     & Max norm 1.0 \\
    Freeze vision tower                   & False \\
    Mixed precision                       & bf16 \\
    
    \midrule
    \end{tabular}
    \captionof{table}{Hyperparameters for GRPO training.}
    \label{appx:grpo_params}
    \end{minipage} \hspace{3mm}
    \begin{minipage}[t]{0.35\linewidth}
  \centering
  \small
    \begin{tabular}{@{} l r @{}} 
    \toprule
    \textbf{Hyperparameter} & \textbf{Value} \\
    \midrule
    Batch size                            & 256 \\
    Max prompt length                     & 2048 \\
    Max response length                   & 2048 \\
    Learning rate                         & 1e-6 \\
    Optimizer                             & AdamW \\
    Scheduler                             & Cosine \\
    Warmup ratio                          & 0.1 \\
    Epochs                                & 2 \\
    Deepspeed config                      & ZeRO Stage 3 \\
    Gradient clipping                     & auto \\
    Freeze vision tower                   & False \\
    Mixed precision                       & bf16 \\
    \midrule
    \end{tabular}
    \captionof{table}{Hyperparameters for SFT training.}
    \label{appx:sft_params}
    \end{minipage}
\end{table*}
\end{document}